\documentclass[11pt]{article}

\usepackage{amsfonts}
\usepackage{amsmath}
\usepackage{amssymb}
\usepackage{amsthm}
\usepackage{float}

\usepackage{graphicx} %
\usepackage{subfigure} 

\usepackage{natbib}

\usepackage{algorithm}
\usepackage{algorithmic}

\usepackage{url}

\usepackage{oursymbols}
\newcommand{\eb}{\epsilon_F}
\renewcommand{\a}{\alpha}
\renewcommand{\b}{\beta}

\newcommand{\X}{{\mathcal X}}
\newcommand{\Y}{{\mathcal Y}}
\newcommand{\F}{{\mathcal F}}

\newcommand{\bpi}{\boldsymbol{\pi}}
\newcommand{\bro}{\boldsymbol{\rho}}
\newcommand{\bxi}{\boldsymbol{\xi}}
\newcommand{\bet}{\boldsymbol{\eta}}

\newcommand{\bsi}{\boldsymbol{\sigma}}
\newcommand{\bnu}{\boldsymbol{\nu}}

\newcommand{\ML}{^{\textrm{{\tiny \textup{ML}}}}}

\newcommand{\prm}{\theta} %
\newcommand{\mix}{\psi} %

\usepackage{color}

\usepackage{bbm}
\renewcommand{\chr}{\boldsymbol{\mathbbm{1}}} %
\newcommand{\pred}[1]{\chr_{\left\{ #1 \right\}}}
\newcommand{\E}{\ensuremath{\text{{\bf\textrm{E}}}}}
\renewcommand{\set}[1]{\left\{ #1 \right\}}
\renewcommand{\vec}{\operatorname{vec}}

\newtheorem{theorem}{Theorem}

\newtheorem{lemma}{Lemma}
\newtheorem{remark}{Remark}

\newcommand{\bepf}{\begin{proof}}
\newcommand{\enpf}{\end{proof}}

\title{On learning parametric-output HMMs}
\author{
Aryeh Kontorovich\\
Department of Computer Science \\
Ben-Gurion University\\
Beer Sheva, Israel
\and
Boaz Nadler\\
Department of Computer Science and applied Mathematics\\
Weizmann Institute of Science\\
Rehovot, Israel
\and
Roi Weiss\\
Department of Computer Science \\
Ben-Gurion University\\
Beer Sheva, Israel
}
\date{}
\begin{document}
\maketitle
\begin{abstract}
We present a novel approach to
learning
an HMM whose outputs are distributed according to a parametric family. 
This is done by {\em decoupling} the learning task into
two steps: 
first estimating the output parameters, 
and then
estimating the 
hidden states
transition probabilities.
The first step is accomplished
by fitting a mixture model to the output 
stationary distribution. 
Given the parameters of this mixture model, 
the second step is formulated as the solution of an easily solvable 
convex quadratic program. 
We provide an error analysis for the estimated 
transition probabilities 
and show they are robust to small perturbations in the estimates 
of the mixture parameters.  
Finally, we 
support our analysis
with some
encouraging
empirical results.
\end{abstract}

\section{Introduction}

Hidden Markov Models (HMM) are a standard tool in the modeling and analysis of time series with a wide variety of applications. 
When the number of hidden states is known, 
the 
standard
method for estimating the HMM parameters
from given observed data is the Baum-Welch algorithm \citep{baumwelch1970}.
The latter is known to suffer from two serious drawbacks:
it tends to converge (i) very slowly and (ii) only to a local maximum.
Indeed, the problem of recovering the parameters of a general HMM
is provably hard, in several distinct senses
\citep{DBLP:journals/ml/AbeW92,Lyngso:2009:CCH:1575125.1575172,Terwijn:2002:LHM:645519.655953}.

In this paper we consider learning parametric-output HMMs with a finite and known number of hidden states, where the output from each hidden state follows a parametric distribution from a given family. A notable example is a Gaussian HMM, where from each state \(x\), the output is a (possibly multivariate) Gaussian,   
\(\mathcal N(\mu_x,\Sigma_x)\), typically with unknown \(\mu_x,\Sigma_x\). 
\hide{
We first consider the case where the output parameters are exactly or approximately known, and later relax this assumption entirely.    
}
\hide{
In addition, there are cases where the output parameters are fixed over time but by some changing
environmental conditions the dynamics of the latent system is changed once in a while 
and we need only to track or learn a new model for the underlying dynamics. 
\hide{
Another
scenario is when we need to classify a sequence of observables according to which latent
dynamical system generated it while fixing the semantics of the latent states. 
}
 Examples include a 
robotic system whose sensors' input/output relations are first calibrated in a laboratory, or a
speech recognition system asking a user to dictate a few specified words
(e.g. a limited supervised initial training phase).
}

\paragraph{Main results.} We propose a novel approach to learning parametric output HMMs, based on the following two insights: 
(i) in an ergodic HMM, the stationary distribution is a mixture of distributions from the parametric family,
and
(ii) given the output parameters, or their approximate values, one
can efficiently recover the corresponding transition probabilities up to
small additive error.

Combining these two insights leads to 
our
{\em decoupling} approach
to learning parametric HMMs.
Rather than attempting,
as in the Baum-Welch algorithm, 
to jointly estimate both the transition probabilities and the output density parameters, 
we instead 
learn each of them separately. First, given one or several long observed sequences, the HMM output parameters are estimated by a general purpose parametric mixture learner, 
such as the Expectation-Maximization (EM) algorithm.
Next, once these parameters are approximately known, 
we learn the hidden state transition probabilities 
by solving a computationally efficient convex quadratic program (QP).

The key idea behind our approach is to treat 
the underlying hidden process as if 
it were sampled independently from 
the Markov chain's stationary distribution, and operate only on the 
empirical distribution of
singletons and consecutive pairs.
Thus we avoid computing the exact likelihood, 
which depends on the full sequence, and obtain considerable gains
in computational efficiency.
Under mild assumptions on the Markov chain and on its output probabilities,
we 
prove 
in Theorem \ref{thm:consist}
that given the exact output probabilities, our estimator for the hidden state transition matrix
is asymptotically consistent.
Additionally, 
this estimator is robust to small perturbations in the output probabilities (Theorems \ref{thm:|pi|}-\ref{thm:Bperturbed}).

Beyond its practical prospects, our proposed approach also sheds light on the theoretical difficulty of the full HMM learning problem: It shows that for parametric-output HMMs the {\em key difficulty is fitting a mixture model}, since once its parameters have been accurately estimated, learning the transition matrix can be cast as a  convex program. 
While learning a general mixture is considered a hard problem, we note that recently much progress has been made under various separation conditions on the mixture components, see e.g. \citet{moitra2010settling, belkin2010polynomial} and references therein.

\paragraph{Related work.}
The problem of estimating HMM parameters from observations has been actively studied since the 1970's, see  
\citet{MR2159833,Rabiner:1990:THM:108235.108253,Roweis:1999:URL:309394.309396}.
While computing the maximum-likelihood estimator for
an HMM is in general computationally intractable, 
under mild conditions,
such an estimator is 
asymptotically consistent and normally distributed, see
\citet{MR1647705,MR1410044,douc2001}. 

In recent years, there 
has been a renewed interest in learning HMMs, 
in particular under various assumptions 
that render the learning problem tractable
\citep{MR1029474,DBLP:conf/colt/HsuKZ09,MR2244426,Siddiqi10,DBLP:conf/colt/Anand12}.
Also, \citet{5773017,5589231} recently suggested Non-negative Matrix Factorization (NNMF) 
approaches for learning HMMs. 
These methods are related to our approach, since with known output probabilities, NNMF\ reduces to a convex 
program
similar to the one considered here. 
Hence, our 
stability and consistency analysis may be relevant to %
NNMF-based approaches as well.  

\paragraph{Paper outline.}
In Section \ref{sec:defs} we present our 
problem setup. The algorithm for learning the HMM appears in Section \ref{sec:algs},
and its statistical analysis in Section \ref{sec:stats}.
Section \ref{sec:simulation} contains some simulation results. The technical details are deferred to the Appendices.

\section{Problem Setup}
\label{sec:defs}

\paragraph{Notation.} 
When $X\in\X$ and $Y\in\Y$ take values in a discrete set we abbreviate 
$P(x)$ for $\Pr(X\!=\!x)$ and $P(y \gn x)$ for $\Pr(Y=y \gn X=x)$. 
When $Y\in\Y$ is continuous-valued, we denote by $P(y \gn x)$ the probability density function of $Y$ given $X$.

For $x,w\in\R^n$, $\diag(x)$  denotes the $n\times n$ diagonal matrix
with  entries $x_i$ on its diagonal, 
$x/w$ is the vector with entries $x_i/w_i$,
and
$\nrm{x}_w^2=\sum_i w_i x_i^2$
is a $w$-weighted $\ell_2$ norm (for $w_i>0$). 
The shorthand $x\lesssim y$  means $x\leq (1+o(1))y$.
Similarly we write $ x \lesssim_P y$ for $x\leq (1+o_P(1))y$.
Finally, for a positive integer $n\in\N$, we write $[n]=\{1,2,\ldots,n\}$.

\paragraph{Hidden Markov Model.}
We consider a discrete-time, discrete-space HMMs 
with $n$ hidden states. The HMM output alphabet, denoted   $\Y, $ 
may be either discrete or continuous. A parametric-output HMM is characterized by a tuple \( (A,\F_n^\prm,P_0) \) where \(A\) is an \(n\times n\) column stochastic matrix, \(P_{0}\) is the distribution of the initial state 
and $\F_n^\prm = (f_{\prm_1},\dots,f_{\prm_n})$ is an ordered tuple 
of {\em parametrized} probability density functions. 
In the sequel we sometimes write $f_i$ instead of $f_{\prm_i}$.

To generate the output sequence of the HMM,\ first an unobserved Markov 
sequence of{ hidden states} 
$x=\{x_t\}_{t=0}^{T-1}$
is generated with the following distribution. 
\beq
P(x) &=& 
P_{0}(x_0)\prod_{t=1}^{T-1} A_{x_t,x_{t-1}},
\eeq
where \(A_{ij} = P(X_t = i\gn X_{t-1} = j)\) are the transition probabilities.
Then, each hidden state $X_t$ independently emits an observation $Y_t\in \Y$ according to the distribution
$P(y_t\gn x_{t})\equiv f_{x_t}(y_t)$. Hence
the output sequence $y=(y_t)_{t=0}^{T-1}$ %
has the conditional probability
$$ P(y\gn x) = \prod_{t=0}^{T-1} P(y_t\gn x_t) = 
\prod_{t=0}^{T-1} f_{x_t}(y_t).$$

\paragraph{The HMM Learning Problem.}
Given one or several HMM output sequences \((Y_t)_{t=0}^{T-1}\), 
the HMM learning problem is to  estimate both the transition matrix $A$ and the parameters of the output distributions $\F_n^\prm$. 

\hide{
\paragraph{Geometric ergodicity.}
Informally, a Markov matrix $A$ is said to be {\em ergodic} if the empirical average of its observations
converges almost surely to the expected value. 
Over finite state spaces,
any ergodic Markov chain is actually {\em geometrically ergodic}:
If $P_t=P_0 A^t $ and $\pi=\lim_{t\to\infty}P_t$ is the stationary distribution,
then
\beqn
\label{eq:geo-erg}
\nrm{P_t-\pi}_1\le 2G\mix^t, \qquad \forall t\in\N,
\eeqn
where 
$G<\infty,\mix\in[0,1)$ are constants determined by $A$. There is a vast literature on the relationship between
various properties of $A$ and the constants $G$ and $\mix$  \citep{kontorovich12,kontram06,MR1410112,kontoyiannis12}.
}

\section{Learning Parametric-Output HMMs}
\label{sec:algs}
The standard approach to learning the parameters of an HMM is to maximize the likelihood 
\beq 
\sum_{x\in[n]^T} 
\!P_{0}(x_0)
P(y_0 \gn x_0)
\prod_{t=1}^{T-1} A_{x_t,x_{t-1}} P(y_t \gn x_t).
\eeq
\hide{
In principle, this likelihood also depends  on the initial state distribution \(P_0\).
However, for \(T\gg 1\), this dependence is very weak, 
and one typically chooses some uniform prior for it, 
maximizing only over
the matrix \(A\) and \(\F_n^\prm\)'s parameters.
}
As discussed in the Introduction, 
this problem is in general computationally hard. 
In practice, neglecting the small effect of the initial distribution
$P_{0}(x_0)$ on the likelihood, \(A\) and \(\F_n^\prm\) are usually 
estimated
via the 
Baum-Welch algorithm, 
which is computationally slow and only guaranteed to converge to a local maximum. 

\subsection{A Decoupling Approach}
In what follows we show that when the output distributions are 
parametric, we can \textit{decouple} the HMM learning task into two steps: 
learning the output parameters
\(\theta_1,\ldots,\theta_n\)
followed by
learning the 
transition probabilities of the HMM. 
Under some mild structural assumptions on the HMM, 
this decoupling implies that the difficulty of learning a parametric-output HMM can be reduced to that of learning a parametric mixture model. Indeed, given (an approximation to) 
$\F_n^\prm$'s parameters, we propose an efficient, single-pass, statistically-consistent algorithm for estimating  
the transition matrix $A$. 

As an example, consider learning a Gaussian HMM with univariate outputs. While the Baum-Welch approach jointly estimates \(n^{2}+2n\) parameters (the matrix $A$ and the parameters $\mu_i,\sigma_i^2$), our decoupling approach first fits a mixture model with only \(3n\) parameters
$(\pi_i,\mu_i,\sigma_i^2)$, and then solves a convex problem for the matrix $A$.  While both problems are in general computationally hard, ours has a 
significantly lower dimensionality for large $n$.

\paragraph{Assumptions. } To recover the matrix \(A\) and the output parameters \(\theta_j\) we make the following assumptions:

(1a) The Markov chain has a unique stationary distribution \(\bpi \) over the $n$ hidden states.
Moreover, 
each hidden state is recurrent with a frequency bounded away from zero:
$\min_k \pi_k\geq a_0$ for some 
constant \(a_0>0\).

(1b) The $n\times n$ transition matrix  \(A\) is geometrically ergodic\footnote{Any finite-state ergodic Markov chain is geometrically ergodic.}: there exists parameters \(G<\infty\)
and $\psi\in [0,1)$ such that from any initial distribution \(P_0\)
\beqn
\label{eq:geo-erg}
\nrm{A^tP_0 -\bpi}_1\le 2G\mix^t, \qquad \forall t\in\N.
\eeqn

(1c) The output parameters of the $n$ states are all distinct: 
$\theta_i\neq \theta_j$ for $i\neq j$. 
In addition, the parametric family is identifiable.

\paragraph{Remarks:} 
Assumption (1a) rules out transient states, whose presence makes it
generally
impossible
to estimate all entries in \(A\) from one or a few long observed sequences. 
Assumption (1b) implies  mixing and is used later on to bound the error and the number of samples needed to learn the matrix \(A.\)
Assumption (1c) is crucial to our approach, which uses the distribution of only single and pairs of consecutive observations. If two states \(i,j\) had same output parameters, it would be impossible to distinguish between them based on single outputs.

\subsection{Learning the output parameters.}
Assumptions (1a,1b) imply that the Markov chain over the hidden states is mixing, and so after only a few time steps, the distribution of \(X_t\) is very close to stationary. 
Assuming for simplicity that already $X_0$ is sampled from the stationary distribution, or alternatively neglecting the first few outputs, this implies that each observable \(Y_t\) is a random realization from the following  {\em parametric mixture model}, 
\begin{equation}
Y\sim \sum_{i=1}^n \pi_i f_{\theta_i}(y).
        \label{eq:p_y}
\end{equation}
Hence, given the output sequence \((Y_t)_{t=0}^{T-1}\) one may estimate the output parameters \(\theta_i\) and the stationary distribution \(\pi_i\)
by fitting a mixture model of the form (\ref{eq:p_y}) to the observations.
This is commonly done via the EM algorithm.

Like its more sophisticated cousin Baum-Welch, the mixture-learning
EM algorithm also suffers from local maxima.
Indeed, from a theoretical viewpoint, learning such a mixture model (i.e. the 
parameters of $\F_n^\prm$) is a non-trivial task considered 
in general to be computationally hard.
\hide{In theory, in
the limit of infinite data, a mixture of Gaussians can be recovered by looking at the far tails, where the density is dominated
by a single component, and thus repeatedly peeling off the components one by one.}
Nonetheless, under various separation assumptions, efficient algorithms with rigorous guarantees have been 
recently proposed (see e.g. \citet{belkin2010polynomial}).\footnote{
Note that the techniques for learning mixtures 
assume
iid data.
However, if these are algorithmically stable --- as such methods typically are --- the iid assumption
can be replaced by strong mixing \citep{mohri2010stability}.
}
Note that while these algorithms have polynomial complexity in sample size and output dimension, they are still exponential in the number of mixture components (i.e., in the number of hidden states of the HMM). Hence, these methods do not imply polynomial learnability of parametric-output HMMs.  

In what follows we assume that using some mixture-learning procedure, the output parameters $\theta_j$ have been estimated with a relatively small error (say  $|\hat\theta_j-\theta_j| = O(1/\sqrt{T})$). 
Furthermore, to allow for cases where \(\theta_j\) were estimated from separate observed sequences of perhaps other HMMs with same output parameters but potentially
different stationary distributions, we do not assume that $\pi_i$  have
been estimated. 

\subsection{Learning the transition matrix $A$}

Next, we describe how to recover the matrix \(A\)
given either exact or approximate knowledge of the HMM output probabilities.
For clarity and completeness, we first 
give an estimation procedure for the
stationary distribution  $\pi$.

\paragraph{Discrete observations.}
\label{subsec:algsDiscrete}
As a warm-up to the case of continuous outputs, we start with  HMMs with a discrete observation space of size  $|\Y| = m$. In this case we can replace $\F_n^\prm$ by
an $m \times n$ column-stochastic matrix $B$ such that $B_{ki} \equiv P(k \gn i)$ is the probability of observing an output $k$ given that the Markov chain is in hidden state $i$. 
In what follows, 
we assume that the number of output states is larger or equal to the number of hidden states, \(m\geq n\), and that the \(m\times n\) matrix \(B\) has full rank \(n\). The latter is the discrete analogue 
of assumption (1c) mentioned above. 

First note that since the matrix $A$ has a stationary 
distribution $\bpi$,
the process $Y_t$ also has a stationary distribution $\bro$, which by analogy to Eq. (\ref{eq:p_y}), is 
\beqn
\label{eq:rhodef}
\bro = B\bpi.
\eeqn
Similarly, the pair $(Y_t,Y_{t+1})$ has a unique stationary distribution
${\bsi}$, given by 
\beqn
\sigma_{k,k'} 
\label{eq:sigdef}
&=&
\sum_{\ell,\ell'\in[n]} \pi_\ell A_{\ell',\ell} B_{k,\ell}B_{k',\ell'}
.
\eeqn

As we shall see below, knowledge of \(\bro\) and $\bsi$ 
suffices to estimate $\bpi$ and $A$. 
Although $\bro$ and $\bsi$ are themselves unknown,
they are easily estimated from a single pass on the data
$(Y_t)_{t=0}^{T-1}$:
\beqn
\hat\rho_k &=& \oo T\sum_{t=0}^{T-1}\pred{y_t=k} ,
\nonumber\\
\label{eq:rhohat}%
\hat\sigma_{k,k'} &=& \oo{T-1}\sum_{t=1}^{T-1}\pred{y_{t-1}=k}\pred{y_t=k'}.
\eeqn
\paragraph{Estimating the stationary distribution $\bpi$.}
The key idea in our approach is to replace the exact, 
but complicated and non-convex likelihood function
by a ``pseudo-likelihood'', which 
treats the hidden state sequence $(X_t)$
as if they were iid draws from 
the unknown stationary distribution \(\bpi\).
The pseudo-likelihood has the advantage of having 
an easily computed global maximum,
which, as we show in
in Section \ref{sec:stats},
yields an asymptotically consistent estimator.
Approximating the $(X_t)$ as iid draws from $\bpi$
means that the $(Y_t)$ are treated as iid draws from
$\bro=B\bpi$.
Thus, 
given a sequence 
$(Y_t)_{t=0}^{T-1}$
the pseudo-likelihood for a vector \(\bpi\) is
\beq
\mathcal L(y_0,\ldots,y_{T-1}\,|\,\bpi)=\prod_{i=0}^{T-1} (B\bpi)_{y_i} = 
        \prod_{k=1}^m (B\bpi)_k^{n_k}
\eeq
where $n_k=\sum_{i=0}^{T-1} \pred{y_t=k} = T \hat\rho_k$. 
Its maximizer is
\begin{equation}
\hat\bpi\ML=
\underset{x_i\geq 0,\,\|x\|_1=1}{\arg\min} 
-\sum_{k=1}^m \hat\rho_k\log(Bx)_k.
        \label{eq:pi_ML}
\end{equation}

Since $-\log(x)$ is convex, \((Bx)_{k}\) is a 
linear combination of the unknown variables \(x_{j}\), and the constraints are all linear, the above is nothing but a {\em convex}
 program, easily solved via standard optimization methods \citep{MR1258086}.
 
However, to facilitate the analysis and to increase the computational efficiency,
we consider the 
asymptotic behavior
of the pseudo-likelihood in (\ref{eq:pi_ML}), 
for
\(T\)  sufficiently large so that $\hat\bro$ is close to $\bro$. 
\hide{To this end, note that if the unknowns in (\ref{eq:pi_ML}) were the \(m\) variables \((Bx)_{k}\),
instead of the $n$ variables $x_i$, then the minimizer would simply be  $(Bx)_k=\hat\rho_k$, for $k=1,\ldots,m$. For $m>n$, this is an over-determined linear system of equations, which in general does not admit a solution 
unless $\hat\bro$ is exactly equal to $\bro$. 
Nonetheless, since for $T$ sufficiently large $\hat\bro$ is close to $\bro$, we expect the solution of (\ref{eq:pi_ML}) 
to satisfy 
$\|Bx-\hat\bro\|_\infty\ll 1$.

This reasoning,  made quantitatively precise in Section \ref{sec:stats},
suggests the following approximation, leading to an even simpler optimization problem. 
}
First, we write
\beq
(Bx)_k=\hat\rho_k \left(1+\frac{(Bx)_k-\hat\rho_k}{\hat\rho_k}\right).
\eeq
Next, assuming that $T\gg 1$ is sufficiently large to ensure $
|(Bx)_k-\hat\rho_k|
\ll \hat\rho_k$, 
we take a second order Taylor expansion of $\log (Bx)_k$ in (\ref{eq:pi_ML}). This gives
\beq
&  & -\sum_{k=1}^n\hat\rho_k\log \hat\rho_k 
 -\sum_{k=1}^n ((Bx)_k-\hat\rho_k) + \nonumber \\
& & \quad +\sum_{k=1}^n \hat\rho_{k}\left(\frac{(Bx)_k-\hat\rho_k}{\hat\rho_k}\right)^2
+O\paren{\frac{\nrm{Bx-\hat\bro}_\infty^3}{\min_j\rho_j^2}}.
\eeq
The first term is independent of $x$, whereas the second term vanishes. 
Thus, we may approximate (\ref{eq:pi_ML}) by the 
{\em quadratic program}
\begin{equation}
\underset{x_i\geq 0,\, \|x\|_1=1}{\argmin}{\|\hat\bro - Bx\|^2_{(1/\hat\bro)}}
        \label{eq:QP_x}
\end{equation}
where $\|x\|_w^2=\sum_k w_kx_k^2$  is a weighted $\ell_2$ norm w.r.t. the weight vector \(w\). 
Eq. (\ref{eq:QP_x}) is also a convex problem, easily solved via standard optimization techniques.
However, let us temporarily ignore the non-negativity constraints $x_i\ge0$ and 
add a Lagrange multiplier for the equality constraint $\sum x_i=1$:
\beqn
\label{eq:QPpi}
\min \oo2 \sum_{k=1}^m \oo{\hat\rho_k}
\Big(\hat\rho_k-\sum_{j=1}^n B_{kj}x_j\Big)^2
-\lambda\Big(\sum_j x_j-1\Big).
\eeqn
Differentiating with respect to $x_i$ yields
\hide{
\beqn
\label{eq:diffx}
\sum_k
\oo{\hat\rho_k}
\Big(
\hat\rho_k-\sum_{j} B_{kj}x_j\Big)
B_{ki}
-\lambda=0
\eeqn
or equivalently}
\begin{equation}
Wx=(1+\lambda)\boldsymbol{1},
        \label{eq:Wx}
\end{equation}
where $W=B\trn\!\diag(1/\hat\bro)B$.
Enforcing the normalization constraint is equivalent to solving for
$x^*=W\inv\boldsymbol{1}$ and normalizing $\hat \bpi=x^*/\nrm{x^*}_1$.
Note that if all entries of 
 $x^*$ are positive, \(\hat\bpi\) is the solution 
of the optimization problem in
(\ref{eq:QP_x}),
and we need not
invoke a QP solver. 
Assumptions (1a,1b) that $\pi_k$
is bounded away from zero and that the chain is mixing imply that for 
sufficiently large $T$, all entries of 
$\hat\bpi$ will be positive 
with high probability, see Section \ref{sec:stats}. 
\hide{
In contrast, if \(x^{*}=W^{-1}{\bf 1}\) has negative entries,
we conclude that \(T\) is too small, and 
there is no point in estimating
the matrix $A$, which requires far more observed data given its larger number of unknowns.
}

\hide{
\begin{remark}
One might naturally attempt to circumvent the complications associated with the 
constrained quadratic program
(\ref{eq:QP_x}) 

by the choice $\hat\bpi=B^+\bro$, where $B^+$ is the Moore-Penrose pseudoinverse. The problem with the latter is that
it typically yields nonstochastic vectors. This can be corrected by 
computing $x^*=B^+\bro$ and
normalizing $\hat\bpi=x^*/\nrm{x^*}_1$, as above (again, we assume
enough observations to guarantee nonnegativity). It is easy to see that 
$\nrm{x^*/\nrm{x^*}_1-\bpi}_2\le2 \nrm{x^*-\bpi}_2$, 
this normalization does not introduce much relative distortion.
\end{remark}
}

\paragraph{Estimating the transition matrix $A$.}
To estimate  $A$,
we consider pairs $(Y_t,Y_{t+1})$ of consecutive observations. 
By definition we have that
for a single pair, 
\beq
P(Y_{t}=k,Y_{t+1}=k') = \sum_{i,j} B_{k'i}B_{kj}A_{ij}P(X_t=j).
\eeq
As above, we  
treat the
\(T-1\) consecutive pairs \((Y_{t},Y_{t+1})\)
as
independent of each other, with the hidden state \(X_{t}\) 
sampled
from the stationary distribution $\bpi$.
When  the output probability  matrix
$B$ and the stationary distribution $\bpi$ are both known, 
the pseudo-likelihood 
is given by
\beq
\calL( y\gn A) = \prod_{(k,k')} 
\Big(
\sum_{ij} B_{k'i}B_{kj}A_{ij}\pi_j \Big)^{n_{kk'}},
\eeq
where $n_{kk'}\!=\!\sum_{t=1}^{T-1}\pred{y_{t-1}=k}\pred{y_{t}=k'}\!=\!(T-1)\hat\sigma_{kk'}$.
The resulting estimator is %
\beqn
\label{eq:QPA}
\underset{
A_{ij}\ge0, \sum_i A_{ij}=1, A\bpi=\bpi
}
{\argmin}{
-\sum \hat\sigma_{kk'}\log\Big( \sum_{ij} C_{ij}^{kk'} A_{ij}\Big)
}
\eeqn
where $C_{ij}^{kk'}=\pi_j B_{kj}B_{k'i}$.
In practice, since $\bpi$ is not known, we use $\hat C_{ij}^{kk'}=\hat\pi_j B_{kj}B_{k'i}$, with
$\hat\bpi$ instead of $\bpi$.
Again, (\ref{eq:QPA}) is a convex program in $A$ and may be solved by standard constrained convex optimization methods.
To obtain a more computationally efficient formulation, 
let us assume that \(\min_{k,k'}\sigma_{k,k'}\geq a_2>0\), and that $\min_{k,k'} T\hat\sigma_{kk'}\gg1$,
so that $|(\hat C A)_{kk'}-\hat\sigma_{kk'}|\ll\hat\sigma_{kk'}$,
where $(\hat CA)_{kk'}=\sum_{ij}\hat C_{ij}^{kk'} A_{ij}$.
Then,
as 
above, 
the approximate minimization problem is
\beqn
\label{eq:QPAappr}
\underset{
A_{ij}\ge0, \sum_i A_{ij}=1, A\hat\bpi=\hat\bpi
}
{\argmin}{
\nrm{\hat\bsi-\hat CA}_{1/\hat\bsi}^2
}.
\eeqn
In contrast to the estimation of $\bpi$, where we could ignore the non-negativity constraints, 
here the constraints $A_{ij}\geq 0$ are essential, since for 
realistic HMMs, some
 entries in $A$ might be strictly zero. Finally, note that if \(\hat\bpi=\bpi\)
 and $\hat\bsi=\bsi$, the true matrix $A$ satisfies $\bsi=CA$ and is 
 the minimizer of (\ref{eq:QPA}).

In summary, given one or more output sequences  \((y_{t})_{t=0}^{T-1}\) and an estimate of \(B,\) we first make a single pass over the data and construct 
the 
 estimators $\hat\bro$ and $\hat\bsi$,
with complexity \(O(T)\). 
Then, the %
stationary distribution $\bpi$ is estimated via (\ref{eq:Wx}), 
and its transition matrix $A$ via  (\ref{eq:QPAappr}). 
To estimate $A$, we first compute the matrix product $\hat C\trn\hat C$, with  $O(n^4 m^2)$ operations.
The resulting QP has size $n^2$, and is thus solvable 
\citep{MR1282714}
in time $O(n^6)$ --- which is dominated by $O(n^4m^2)$ since $m\ge n$ by assumption.
Hence, the overall time complexity of estimating \(A\) is  \(O(T+n^{4}m^2)\).
\paragraph{Extension to continuous observations.}
\label{subsec:algsParam}
We now extend the above results to the case of continuous outputs distributed according to a known parametric family. Recall that in this case, each hidden state $i\in [n]$ has an associated output probability density $f_{\prm_i}(y)$. 
As with discrete observations, we assume that an approximation $(\hat{\prm}_1,\dots,\hat{\prm}_n)$ to $f_i$'s parameters is given
and use it to
construct estimates of $\bpi$ and $A$. 

To this end, we seek analogues of 
(\ref{eq:rhodef}) and (\ref{eq:sigdef}), which relate the observable quantities to the latent ones. 
This will enable us to construct the appropriate empirical estimates and the corresponding quadratic programs, 
whose solutions will 
be
our estimators $\hat{\bpi}$ and $\hat{A}$. 
To handle 
infinite output alphabets,
we map each observation \(y\) to an \(n\)-dimensional vector 
\(\varphi(y)=(f_{\theta_1}(y),\ldots,f_{\theta_n}(y))\), whose entries are  the likelihood of $y$ from each of the underlying hidden states. As shown below, this allows us to reduce the problem to a discrete ``observation'' 
space which can be solved by the methods introduced in the previous subsection.

\hide{
For simplicity, in the rest of this section we assume that the HMM is started with the stationary distribution $\pi$. As discussed in section \ref{sec:stats}, this assumption 
is not at all restrictive.
}

\paragraph{Estimating the stationary distribution $\bpi$.}
To obtain an analogue of (\ref{eq:rhodef}), we define the vector $\bxi \in \R^n$, and matrix $K\in\R^{n \times n}$, which
will play the role of $\bro$ and $B$ for discrete output alphabets.
The vector \(\bxi\) is defined as \(\bxi=\E[\varphi(Y)]\), 
or more explicitly,
\beq
\xi_k & \equiv &  \E[f_k(Y)]  =  \sum_{j=1}^{n} \pi_j \int_\Y f_k(y) P(y \gn j) dy.
\eeq
Similarly, the \((i,j)\) entry of $K$ is given by
\begin{equation}
K_{ij} \equiv \E[f_i(Y) \gn X = j ] = \int_{\mathcal Y} f_i(y) P(y \gn j) dy.
        \label{eq:K_def}
\end{equation}
With these definitions we have, as in Eq.  (\ref{eq:rhodef}),
\beqn
\label{eq:xi}
\bxi = K \bpi.
\eeqn
Thus, given an observed sequence 
$(y_t)_{t=0}^{T-1}$
we construct the empirical estimate
\beqn
\label{eq:xihat}
\hat{\xi}_k = \oo{T} \sum_{t=0}^{T-1} f_k(y_t),
\eeqn
and consequently solve the QP
\beqn
\label{eq:piParam}
\hat{\bpi} = \underset{\nrm{x}_1 = 1, x \geq 0}{\arg\!\min}
 \nrm{\hat{\bxi} - Kx}_{1/\hat{\bxi}}^2.
\eeqn
In analogy to the discrete case, 
we assume \(\rank(K)=n\) so (\ref{eq:piParam})
has a unique solution. Its asymptotic consistency and accuracy are discussed 
in Section \ref{sec:stats}.
\paragraph{Estimating the transition matrix $A$.}
Next, following the same paradigm we obtain an analogue of (\ref{eq:sigdef}).  
 Bayes rule implies that for stationary chains,
\beqn
\label{eq:P(k|y)}
P(k \gn Y) = \frac{f_k(Y) \pi_k}{\sum_{l=1}^{n}{f_l(Y) \pi_l}}.
\eeqn
We define the matrices
$\bet\in \R^{n \times n}$ and $F\in \R^{n \times n}$ 
(analogues of $\bsi$ and $B$)
as follows.
Let $Y$ and $Y'$ be two {\em consecutive} observations of the HMM, then
\begin{eqnarray}
\eta_{kk'} & \!\!\equiv\!\! & \E \left[P(k \gn Y) P(k' \gn Y') \right] \nonumber\\
F_{kj} &\!\!\equiv\!\!& \E[P(k \gn Y) \gn j] \! = \!\! \int_{\mathcal Y} P(k \gn y) P(y \gn j) dy. \label{eq:def_F}
\end{eqnarray} 
\hide{
\beq
\eta_{kk'} & \equiv & \E \left[P(k \gn Y) P(k' \gn Y') \right],
\eeq
and %
\beq
F_{kj} &\equiv& \E[P(k \gn Y) \gn j] = \int_{\mathcal Y} P(k \gn y) P(y \gn j) dy.
\eeq
}
A simple calculation shows that, as in (\ref{eq:sigdef}),
\beqn
\label{eq:Etakk}
\eta_{kk'}  =  \sum_{i,j = 1}^n F_{k'i}F_{kj}A_{ij}\pi_j.
\eeqn

Since
here
$F$
plays the role of $B,$
we may call it
an {\em effective} observation matrix.
This 
suggests estimating $A$ with the same tools used in the discrete case.
Thus, given an observed  sequence \((y_{t})_{t=0}^{T-1}\) 
\hide{
and assuming a known \(\bpi\), the natural estimator for $\bet$ is
\beq
\oo{T-1} \sum_{t=1}^{T-1} P(k \gn y_{t-1}) P(k' \gn y_{t}).
\eeq
In practice, since we only have an estimate \(\hat\bpi\) of  ${\bpi}$, }
we construct an  empirical estimate $\hat{\bet}$  by
\beqn
\label{eq:etahat}
\hat{\eta}_{k k'} & = &\oo{T-1} \sum_{t=1}^{T-1} \hat{P}(k \gn y_{t-1}) \hat{P}(k' \gn y_{t}),
\eeqn
where $\hat{P}$ is given by (\ref{eq:P(k|y)}) but with $\pi$ replaced by $\hat{\pi}$.
\hide{
\beq
\hat{P}(i \gn y) = \frac{P(y \gn i) \hat{\pi}_i}{\sum_{k=1}^{n}{P(y \gn k) \hat{\pi}_k}}.
\eeq
}
Consequently we solve the following QP
\beqn
\label{eq:QPAParam}
\hat{A} = \underset{A_{ij} \geq 0, \sum_{i}A_{ij}=1,A\hat{\bpi}=\hat{\bpi}}{\arg\!\min}
 \nrm{\hat{\bet} - (\hat{C}A)}_{1/\hat{\bet}}^2,
\eeqn
where  $\hat{C}_{ij}^{kk'}=\hat{\pi}_j F_{kj} F_{k' i}$ and $(\hat{C}A)_{kk'}=\sum_{ij} \hat{C}_{ij}^{kk'}A_{ij}$. As for the matrix \(B\) in the discrete case, to ensure a unique solution to Eq. (\ref{eq:QPAParam})
we assume \(\rank(F)=n.\) 

\hide{
Note that $\hat{\xi}_{k k'}$ is $2/(T-1)$-Lipschitz so it obeys the concentration lemma in \ref{}.
}

\begin{remark}
\label{rem:Kestimation}
Instead of (\ref{eq:Etakk}), we could estimate 
$\eta'_{k,k'}  \equiv  \E[f_k(Y) f_{k'}(Y')]$, from which $A$ can also be recovered, since 
\beq
\label{eq:Xikktag}
\eta'_{k,k'}  %
                 =  \sum_{i,j = 1}^n  K_{k'i} K_{kj} A_{ij} \pi_j  .
\eeq
This has the advantage that for many distributions the matrix $K$ can be cast in a closed analytic form. For example in the Gaussian case, while $F$ needs to be calculated numerically, we have
\beq
K_{ij} = \oo{\sqrt{2\pi}} \oo {\sqrt{\sigma_i^2 + \sigma_j^2}}
\exp
\left(
-\oo{2} \frac{(\mu_i - \mu_j)^2}{\sigma_i^2 + \sigma_j^2}
\right).
\eeq
Additionally, $K$ does not depend on the stationary distribution. The drawback is that in principle, and as simulations suggest, accurately estimating $\bet'$ may require many more samples, see Appendix for details.  
\end{remark}

In summary, given approximate output parameters $(\hat{\prm}_1,\dots,\hat{\prm}_n)$, we first calculate the $n \times n$ matrix $K$. 
Next, we construct the vector $\hat{\bxi}$ by a single pass over the data  
$(Y_t)_{t = 0}^{T-1}$. Then the stationary distribution $\bpi$ is estimated via (\ref{eq:piParam}). Given  $\hat{\bpi}$, we  calculate the $n \times n$ matrix $F$, construct the empirical estimate $\hat{\bet}$, and estimate $A$ via (\ref{eq:QPAParam}). As in the discrete observation case, the time complexity of this scheme is $O(T + n^6)$ with additional terms for calculating $K$ and $F$.

\section{Error analysis}

\newcommand{\bv}{{\bf v}}
\newcommand{\bu}{{\bf u}}
\label{sec:stats}
First, we study the statistical properties of our estimators under the assumption that the output parameters, $({\prm}_1,\dots,{\prm}_n)$ in the continuous case, or the matrix $B$ in the discrete case, are known {\em exactly}. Later on  
we show that our estimators are stable to perturbations in these parameters. 
For simplicity, throughout this section we assume that the initial hidden state $X_0$ is sampled
from the stationary distribution $\bpi$. This assumption is not essential
and 
omitting it
would not qualitatively change our results. All proofs are deferred to the Appendices.

To provide bounds on the error and required sample size we make the following additional assumptions: 

(2a) In the discrete case, there exists an \(a_{1}>0\) such that  \(\min_j \rho_j\geq a_1\).

(2b) In the continuous case, all $f_{\prm_i}$ are bounded:
$$ \max_{i\in [n]} \sup_{y\in\R} f_{\prm_i}(y)\le L<\infty.$$

Finally, for ease of notation we define
\beq
g_\mix \equiv \frac{2G}{1-\mix}.
\eeq

\paragraph{Asymptotic Strong Consistency.}

Our first result shows that with perfectly known output probabilities, as \(T\to\infty\), our estimates \(\hat\bpi,\hat A\) are strongly consistent.

\begin{theorem}
\label{thm:consist}
Let $(Y_t)_{t=0}^{T-1}$ be an observed sequence of an HMM, whose Markov chain satisfies Assumptions (1a,1b). Assume \(\rank(B)=n\) in the discrete case, or 
$\rank(F)=\rank(K)=n$ in the continuous case. Then, both estimators, 
$\hat\bpi$ of (\ref{eq:Wx}) and $\hat A$ of (\ref{eq:QPAappr}) in the discrete case, or (\ref{eq:piParam}) and (\ref{eq:QPAParam}) in the continuous case,
are 
asymptotically strongly consistent. Namely, as $T\to\infty$, 
with probability one,%
\beq
\hat\bpi \to\bpi\quad\mbox{and}\quad 
\hat A \to A.
\eeq
\end{theorem}

\paragraph{Error analysis for the stationary distribution $\bpi$.}
\label{sec:err-pi}

Recall that to estimate  \(\bpi\) in the discrete case, we argued that for sufficiently large sample size \(T\), the positivity constraints 
can be ignored, which amounts to solving an \(n\times n\) system of linear equations, Eq. (\ref{eq:Wx}). The following theorem provides both a lower bound on the required sample size \(T\) for this condition to hold with high probability, as well as error bounds on the difference \(\hat\bpi-\bpi\). 
\hide{
A  theoretical question is how large \(T\) must be
so that indeed the solution of (\ref{eq:Wx}) is strictly positive in all its coordinates, 
with high probability. 
A second question is how far is $\hat\bpi$ from $\bpi$.
}
\hide{ both in the discrete and continuous cases.}
\hide{These questions are addressed in the following theorem.}
\begin{theorem}
\label{thm:|pi|}
{\bf Discrete case}: Let $\hat\bro$ be given by (\ref{eq:rhohat}), and $\hat\bpi$ be the solution of (\ref{eq:Wx}).
Let \(\tilde{B}=\diag(1/\sqrt{\bro})B\), and \(\sigma_1(\tilde B)\)
be its smallest singular value. 
Under Assumption (2a), a sequence of length
\beqn 
\label{eq:Tpi}
T\gtrsim 
\frac{g_\mix \sqrt{\log n}}{a_0 a_1 \sigma_1(\tilde B)}
,
\eeqn
is sufficient to ensure that
with high probability,
all entries in $\hat\bpi$ 
are strictly positive.
Furthermore, as $T\to\infty$,
\beqn
 \nrm{\hat\bpi-\bpi}_2 \lesssim_P
\hide{ \nrm{\hat\bpi-\bpi}_2 &\le&
(1+o_P(1))}
\sqrt{\frac{ g_\mix^2}{T a_1^2 \sigma_1^2(\tilde B) }}
.
        \label{eq:Error_pi}
\eeqn
\end{theorem}
Next we consider the errors in the estimate $\hat\bpi$ for the continuous observations case. For simplicity, instead of analyzing the quadratic program (\ref{eq:piParam}) with a weighted \(\ell_{2}\) norm, 
we consider 
the following quadratic program, whose solution is also asymptotically consistent:
\beqn
\label{eq:QPpiParam}
\min_{x \geq 0,\sum_i x_{i}=1} \|\hat\bxi- K x \|^2_2.
\eeqn
This allows for a cleaner analysis, without changing the qualitative flavor of the results.
\begin{theorem}
\label{thm:|pi|Par}
{\bf Continuous case}: Let $\hat\bxi$ be given by (\ref{eq:xihat}),  $\hat\bpi$ be the solution of (\ref{eq:piParam}), and \(\tilde{K}=\diag(1/\sqrt{\bxi})K.\) Under Assumption (2b), as $T\to\infty$,
\beqn
 \nrm{\hat\bpi-\bpi}_2 \lesssim_P
\hide{ \nrm{\hat\bpi-\bpi}_2 &\le&
(1+o_P(1))
}
\sqrt\frac{(n^3 \ln n)  g_\mix^2 L^4 }{ T \sigma_1^4(\tilde K) }
,
        \label{eq:Error_pi_param}
\eeqn
\end{theorem}
\paragraph{Error Analysis for the  Matrix $A$.}

\hide{
 Since both $B$ and $F$ are stochastic we can treat them both simultaneously. Therefore we write $F$ for both $B$ and $F$ and $\bet$ for both $\bsi$ and $\bet$.
}
 Again, for simplicity, instead of analyzing the quadratic programs (\ref{eq:QPAappr}) and (\ref{eq:QPAParam}) with a weighted \(\ell_{2}\) norm, 
we consider 
the following quadratic programs, whose solutions are also asymptotically consistent for $\hat{\bnu} \in \{\hat{\bsi},\hat{\bet}\}$:
\beqn
\label{eq:|CA|}
\min_{A_{ij}\geq 0,\sum_i A_{ij}=1} \|\hat\bnu-\hat C A \|^2_2 .
\eeqn
Note that this QP is applicable even if $\nu_{kk'}=0$ for some $k,k'$, which implies that
$\hat\nu_{kk'}=0$ as well.

\begin{theorem}
{\bf Discrete case.}
\label{thm:main}
Let $\hat A$ be the solution of (\ref{eq:|CA|}) with $\hat{\bnu} = \hat{\bsi}$ given in (\ref{eq:rhohat}).
Then, as $T\to\infty$,
\begin{equation}  
        \label{eq:ERROR_A_HATd}
\nrm{\hat A-A}_F \lesssim_P 
\hide{\nrm{\hat A-A}_F \le 
(1 + o_P(1))}
\sqrt{\frac{n^{3} g_\mix^2}{T a_0^4 a_1^2 \sigma_1^{10}(B)}}
\end{equation}
and thus an observed sequence length
\beqn
\label{eq:TAd}
T
\gtrsim  
\frac{n^3 g_\mix^2}{ a_0^4 a_1^2\sigma_1^{10}(B)}
\eeqn
suffices for accurate estimation.
\end{theorem}

\begin{theorem}
{\bf Continuous case.}
\label{thm:mainPar}
Let $\hat A$ be the solution of (\ref{eq:|CA|}) with $\hat{\bnu} = \hat{\bet}$ given in (\ref{eq:etahat}).
Then, as $T\to\infty$,
\begin{equation}  
        \label{eq:ERROR_A_HATc}
\nrm{\hat A-A}_F \lesssim_P 
\hide{
\nrm{\hat A-A}_F \le 
(1 + o_P(1))
}
\sqrt{\frac{(n^{7} \ln n) g_\mix^2 L^4 }{T a_0^6\sigma_1^{8}(F) \sigma_1^{4}(K)}}
\end{equation}
and thus an observed sequence length
\beqn
\label{eq:TAc}
T
\gtrsim  
\frac{(n^7 \ln n) g_\mix^2 L^4}{a_0^4 \sigma_1^{8}(F) \sigma_1^{4}(K)}
\eeqn
suffices for accurate estimation.
\end{theorem}

\begin{figure}[ht]
\centering
{\includegraphics[width=0.53\textwidth]{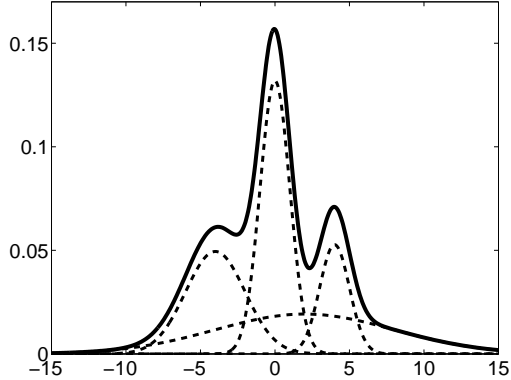}}
\caption{The mixture and its components.}
\label{fig:Mixture}
\end{figure} 

{\bf Remarks.} Note the key role of the smallest singular value \(\sigma_1\), in the error bounds in the theorems above:
Two hidden states with very similar output probabilities drive $\sigma_1$ to zero, thus 
requiring many more observations to resolve the properties of the underlying hidden sequence.

\paragraph{Inaccuracies in the output parameters.}
In practice we only have  approximate output parameters, found for example, via an EM algorithm. For simplicity, we study the effect of such inaccuracies only in the continuous case. Similar results hold in the discrete case. To this end,  assume the errors in the matrices \(K\) and $F$ of Eqs. (\ref{eq:K_def}) and (\ref{eq:def_F}) are of the form
\beqn
\label{eq:pert}
\tilde{K} = K + \epsilon L Q,\quad\quad \tilde{F} = F + \epsilon  P,
\eeqn
with $\nrm{Q}_F ,\nrm{P}_F \leq 1$. The following theorem shows our estimators are \textit{stable} w.r.t. errors in the estimated output parameters.
Note that if $K,F$ are estimated by a sequence of length \(T\), 
then typically $\epsilon = O(T^{-1/2})$.

\begin{theorem}
\label{thm:Bperturbed}
Given an error of $\epsilon$ in the output parameters as in Eq. \eqref{eq:pert}, 
the estimators given in Theorems \ref{thm:|pi|Par} and \ref{thm:mainPar}, incur an additional error of at most 
\beqn
O 
\left(
\frac{n^{r}\epsilon}{a_0^2\sigma_1^4}
\right),
\eeqn
with $r=1$ for estimating $\bpi$, and $r=\frac{3}{2}$ for estimating $A$, and where  $\sigma_1$ is the smallest singular value of $K/L^2$ when estimating $\bpi$, and of $F$ when estimating $A$.
\end{theorem}
\hide{
Theorem \ref{thm:Bperturbed} thus shows our estimators to be \textit{stable} with respect to errors in the estimated output parameters.
Note that if $K,F$ are estimated using a sequence of length \(T\), 
then typically $\epsilon = O(T^{-1/2})$.
}

\hide{
\begin{figure*}[t]
\centering
\subfigure[$\E\|\hat A-A\|_F^2$ vs. $T$.]
{\includegraphics[width=0.215\textwidth]{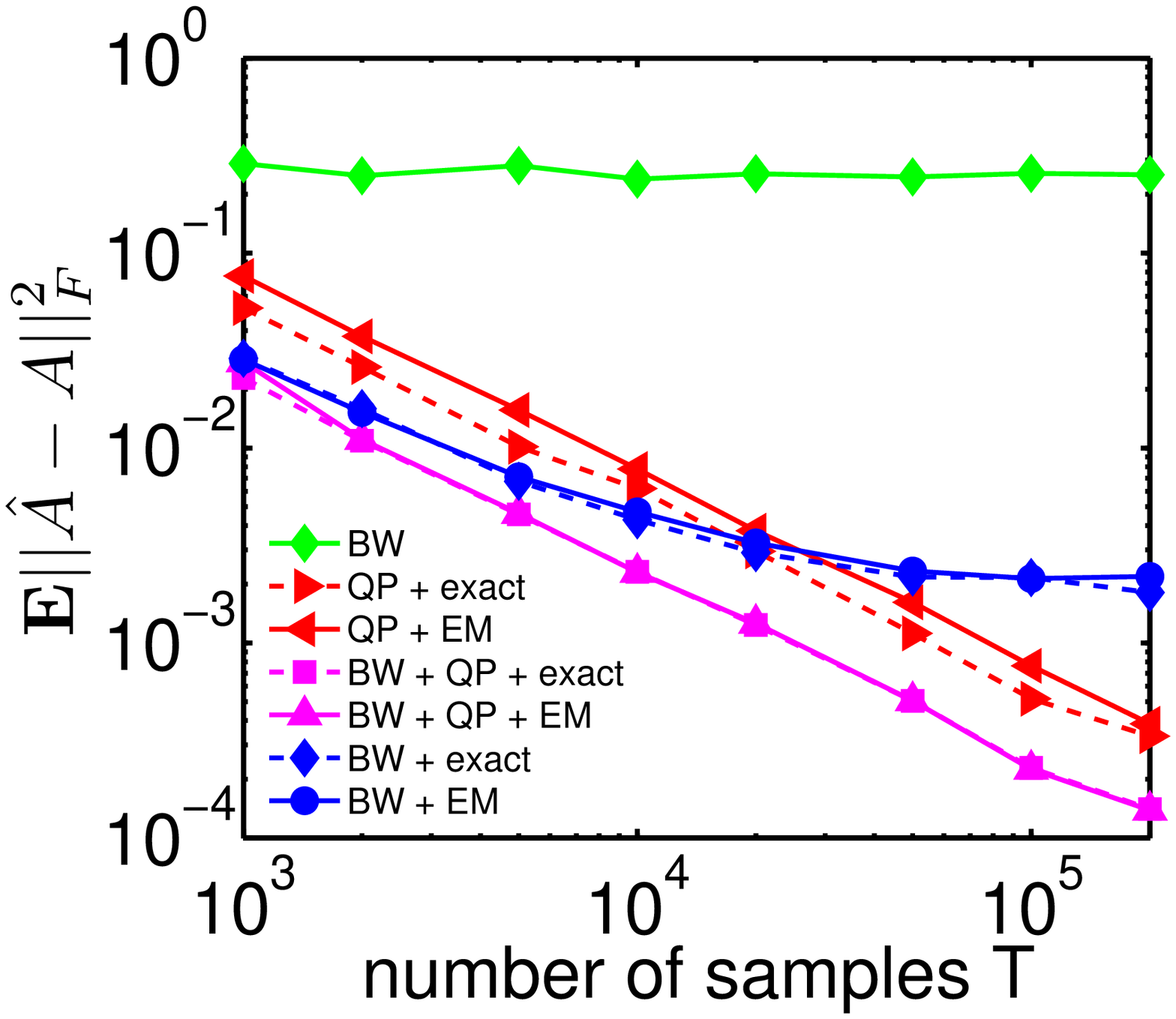}\label{fig:ErrVsT}}
\hspace{0.\textwidth}
\subfigure[running times vs. $T$.]
{\includegraphics[width= .235\textwidth]{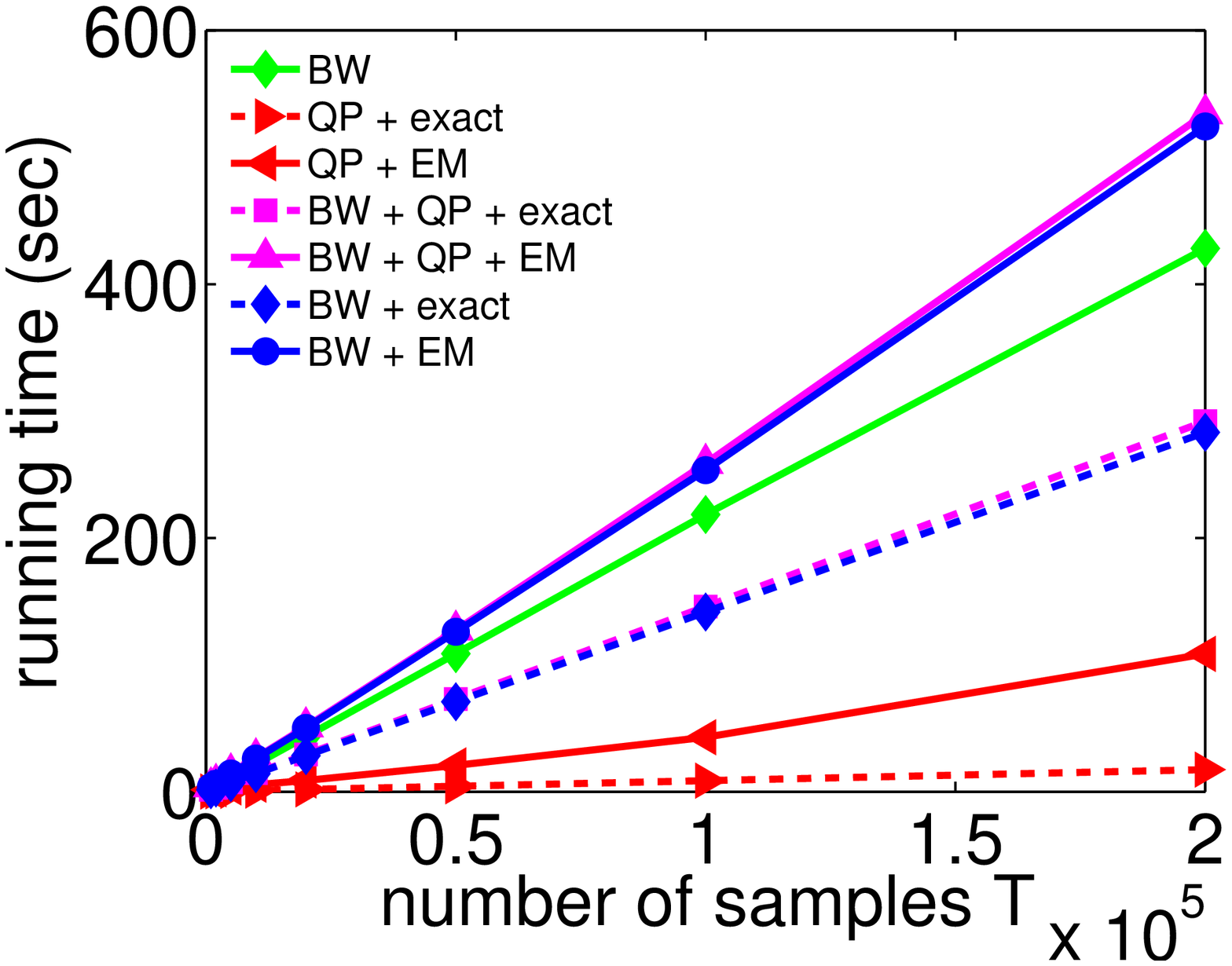}\label{fig:RunningVsT}}
\centering
\subfigure[$\E\|\hat A-A\|_F^2$ vs. number of BW iterations.]
{\includegraphics[width=0.225\textwidth]{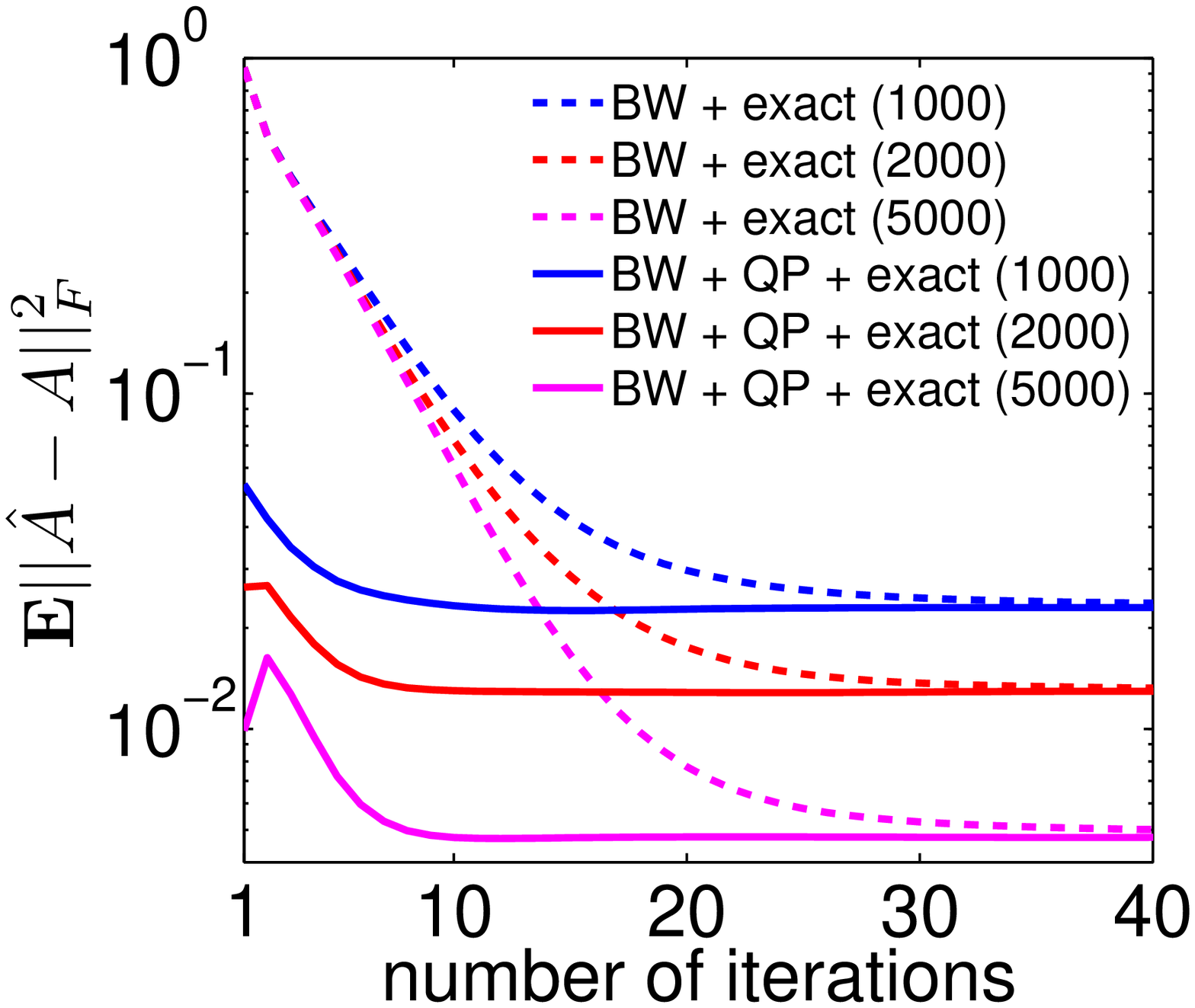}\label{fig:ErrVsIter}}
\hspace{0.\textwidth}
\subfigure[$\E\|\hat A-A\|_F^2$ vs. number of BW iterations.]
{\includegraphics[width=0.225\textwidth]{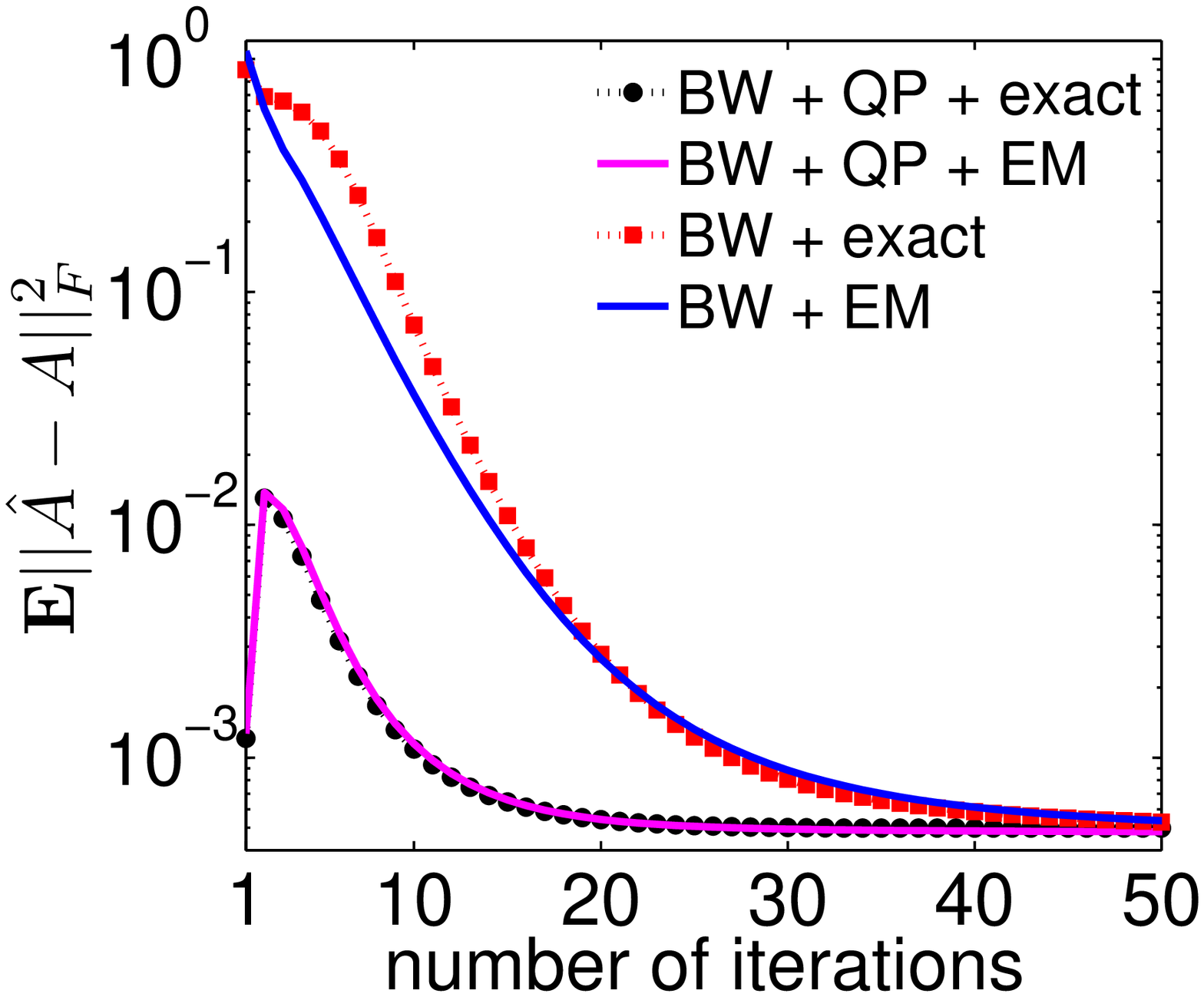}\label{fig:ErrVsIterAll}}
\caption{Simulation results}
\end{figure*}
}

\section{Simulation Results}
\label{sec:simulation}

We illustrate our algorithm by some simulation results, executed in MATLAB with the help of the
HMM and EM toolboxes\footnote{Available at \url{http://www.cs.ubc.ca/~murphyk} and \url{http://www.mathworks.com/} (under EM\_GM\_Fast).}.
We consider a toy example with $n=4$ hidden states, whose outputs are univariate Gaussians, $\mathcal{N}(\mu_i,\sigma_i^2)$, 
with $A$, $\F_n^\prm$ and $\bpi$ given by
{
\beq
A \!&\!=\!\left(\begin{array}{cccc}
  0.7 & 0.0 & 0.2 & 0.5 \\
  0.2 & 0.6 & 0.2 & 0.0 \\
  0.1 & 0.2 & 0.6 & 0.0 \\
  0.0 & 0.2 & 0.0 & 0.5
\end{array}
\right)  , 
\quad 
\begin{array}{rcl}
f_1 &=& \mathcal{N}(-4,4)\\
f_2 &=& \mathcal{N}(0,1)\\
f_3 &=& \mathcal{N}(2,36)\\
f_4 &=& \mathcal{N}(4,1)
\end{array}\\
& \bpi\trn =   (0.3529,    0.2941,    0.2353,    0.1176).
\eeq
}
Fig. \ref{fig:Mixture} shows the mixture and its four components.

To estimate \(A\) we considered the following methods:

\begin{center}
\begin{tabular}{|c|l|l|l|}
        \hline
 &  method    &   initial $\prm$  & initial $A$   \\
        \hline
1   & BW & random & random    \\
2   & none & exactly known &  QP \\
3   & none & EM &  QP \\  
4   & BW   & exactly known &  QP \\
5   & BW   & EM  &  QP      \\
6   & BW   & exactly known & random  \\
7   & BW   & EM  & random  \\
        \hline
\end{tabular}
\end{center}
\hide{
\begin{itemize}
\item[(1)] the BW algorithm with a random initial guess.
\item[(2)] our QP with exactly known emission parameters.
\item[(3)] our QP with emission parameters estimated by the EM algorithm.
\item[(4)] BW with known emission parameters and initial guess of \(A \) obtained by our QP in item (2). 
\item[(5)] BW with emissions parameters estimated by EM and initial guess of \(A\)  by our QP in item (3).
\item[(6)] BW with known emission parameters and random guess of \(A\). 
\item[(7)] BW with initial parameters obtained by EM and random guess of \(A\).
\end{itemize}
}

Fig. \ref{fig:ErrVsT} (left) shows on a logarithmic scale ${\E\|\hat A-A\|_F^2}$ vs. sample size \(T\),   averaged over 100 independent realizations. 
Fig. \ref{fig:ErrVsT} (right) shows the running time as a function of \(T\). In these two figures, the number of iterations of the BW step was set to 20.

 Fig. \ref{fig:ErrVsIterAll} (left) shows the convergence of ${\E\|\hat A-A\|_F^2}$ as a function of the number of BW iterations, with known output parameters, but either with or without the QP results. Fig. \ref{fig:ErrVsIterAll} (right) gives ${\E\|\hat A-A\|_F^2}$ as a function of the number of BW iterations for both known and EM-estimated output parameters with $10^5$ samples.

\begin{figure*}[t]
\centering
\includegraphics[width=.42\textwidth]{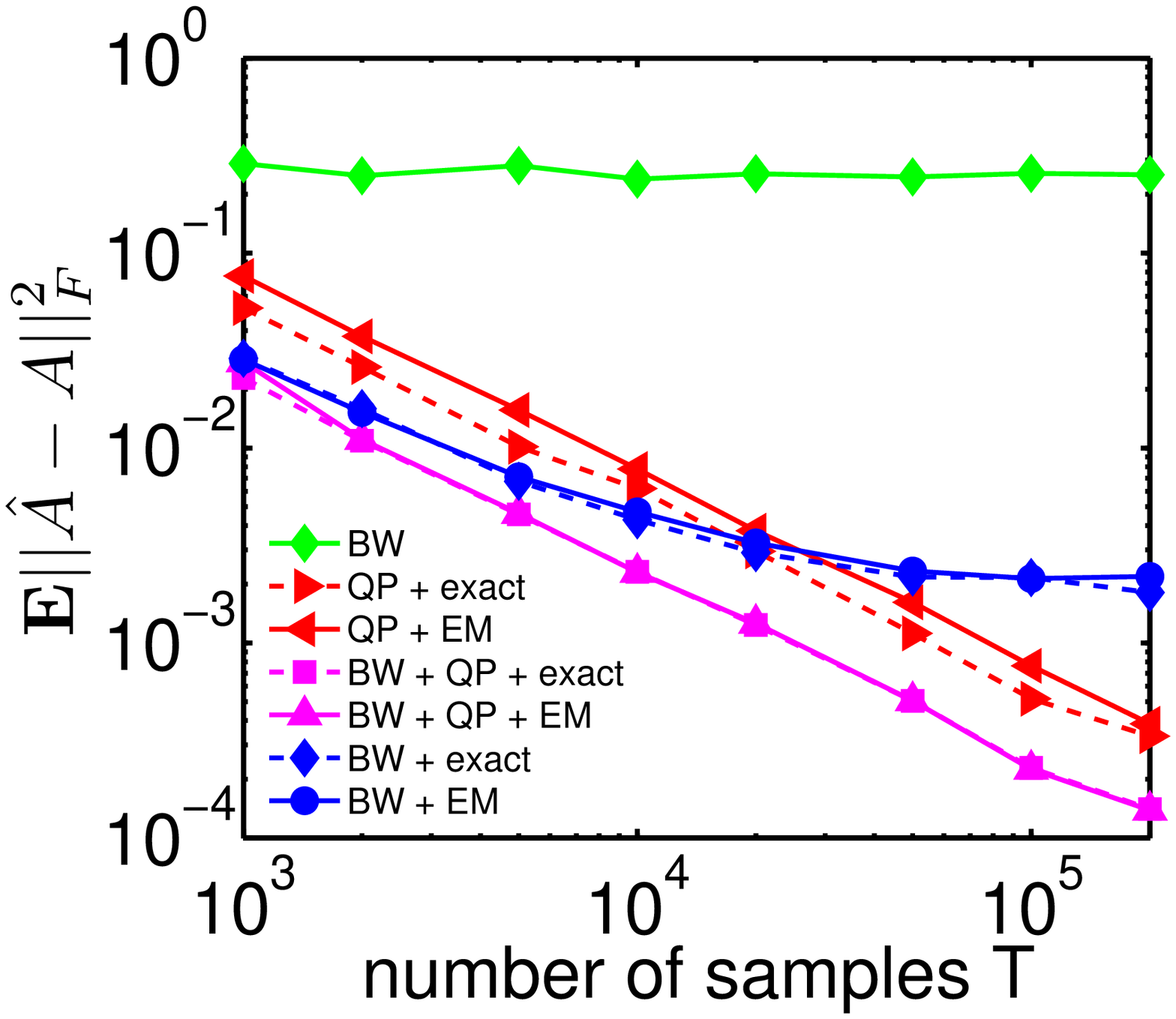}%
\hspace{0.1\textwidth}
{\includegraphics[width= .46\textwidth]{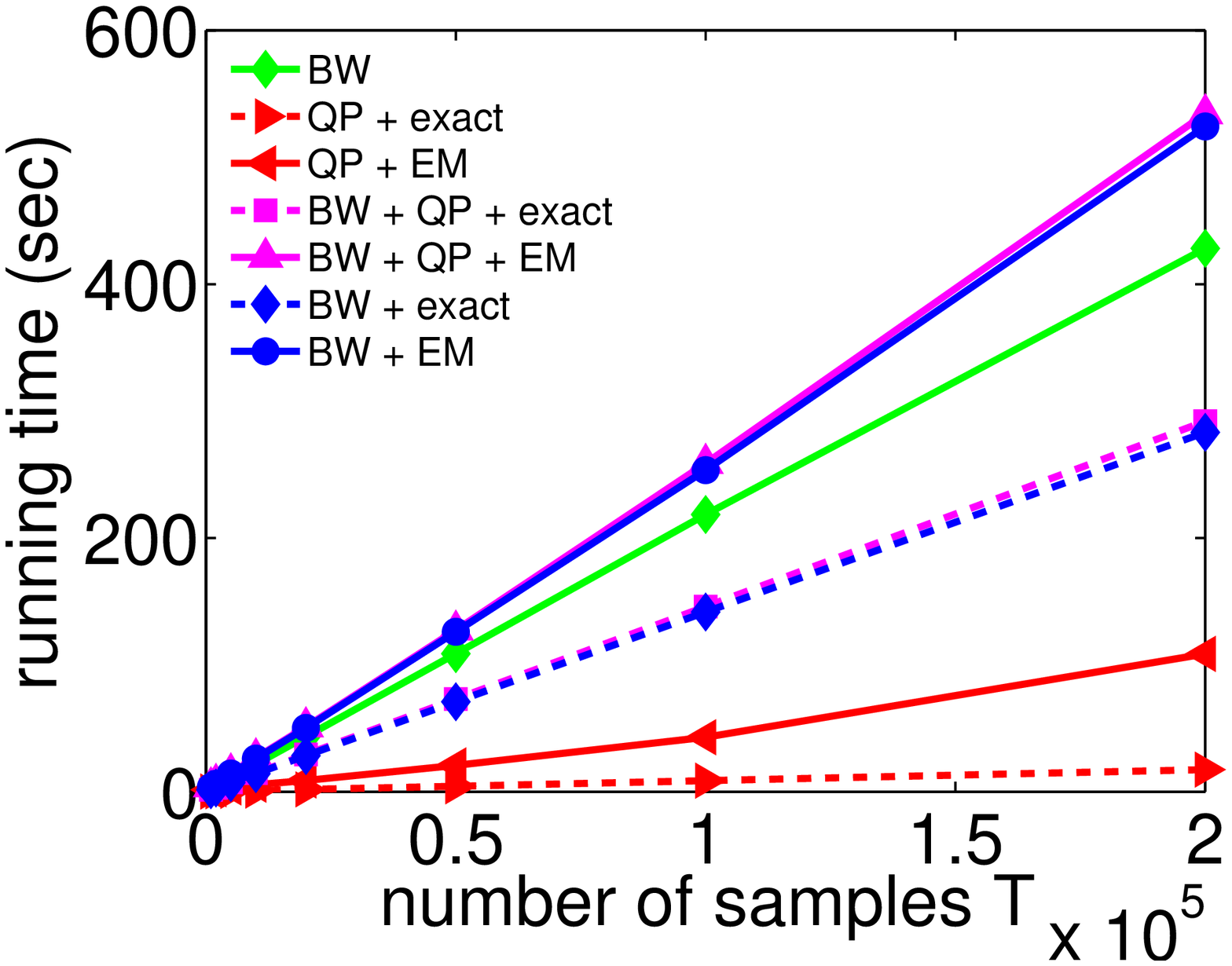}}%
\caption{Average Error $\E\|\hat A-A\|_F^2$ and runtime comparison of different algorithms vs. sample size \(T\).}
\label{fig:ErrVsT}
\end{figure*}

\begin{figure}
\centering
{\includegraphics[width=0.4\textwidth]{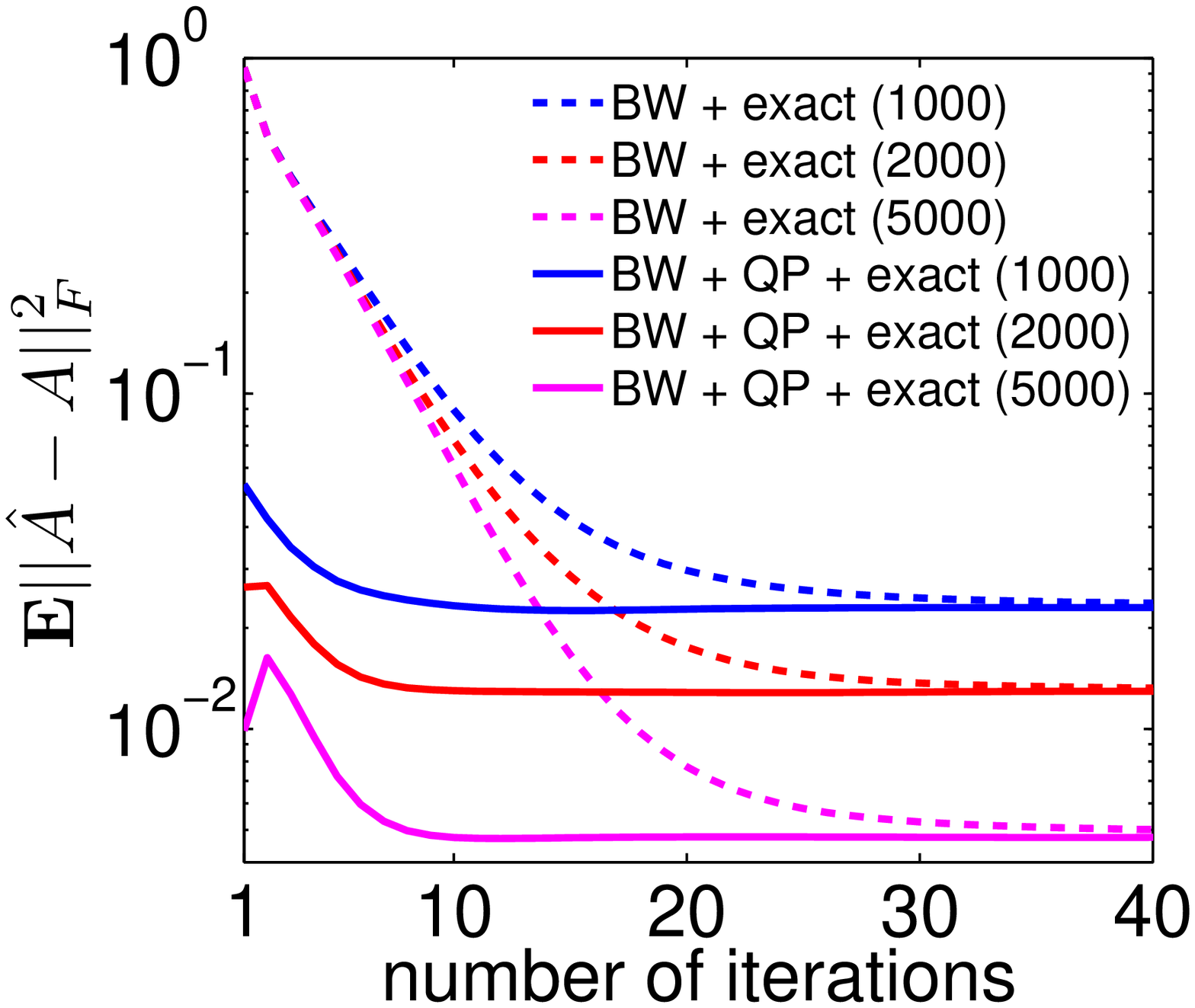}}
\hspace{0.1\textwidth}
{\includegraphics[width=0.4\textwidth]{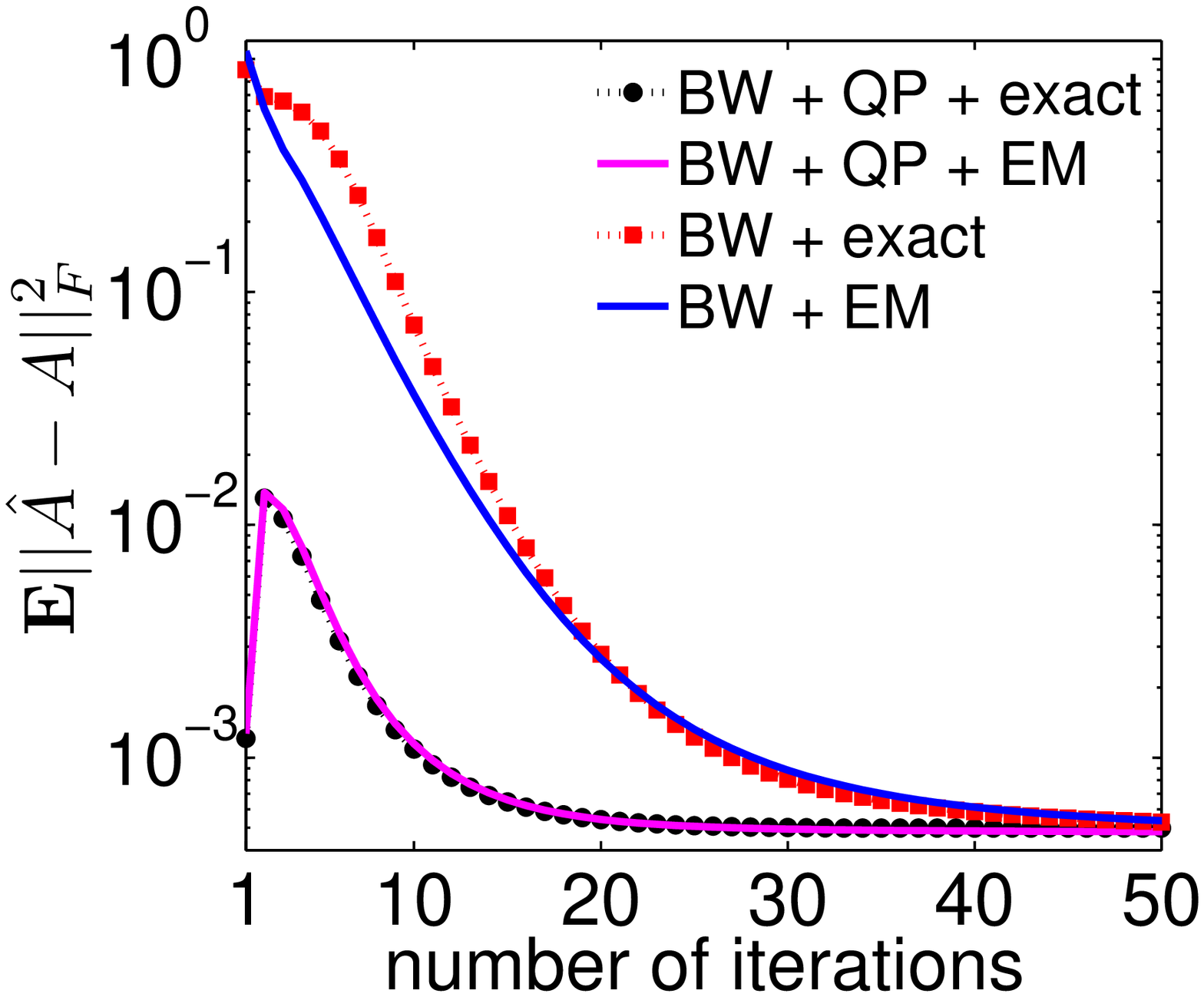}}
\caption{Convergence of the BW iterations.}
\label{fig:ErrVsIterAll}
\end{figure}
The simulation results highlight the following points: 
(i) BW with a random guess of both \(A\) and the parameters \(\theta_j=(\mu_j,\sigma_j^2)\) is useless if run for only 20 iterations. It often requires hundreds of iterations to converge, in some cases to a poor inaccurate solution (results not shows due to lack of space); (ii) For a small number of samples the accuracy of QP+EM (method 3) is comparable to BW+EM (method 5) but requires only a fraction of the computation time. (iii) When the number of samples becomes large, the QP+EM is not only faster, but (surprisingly) 
also more accurate than BW+EM.
As Fig. \ref{fig:ErrVsIterAll} suggests, this is due to the slow convergence of the BW algorithm, which requires more than 20 iterations for convergence.
(iv) Starting the BW iterations with \((\mu_i,\sigma_i^2)\) estimated by EM and $A$ estimated by QP as its initial values significantly accelerated the convergence giving a superior accuracy after only 20 iterations. These results show the (well known) importance of initializing the BW algorithm with sufficiently accurate starting values. Our QP approach provides such an initial value for \(A\) by a computationally fast algorithm.   

\bibliography{ourbib}

\begin{thebibliography}{25}
\providecommand{\natexlab}[1]{#1}
\providecommand{\url}[1]{\texttt{#1}}
\expandafter\ifx\csname urlstyle\endcsname\relax
  \providecommand{\doi}[1]{doi: #1}\else
  \providecommand{\doi}{doi: \begingroup \urlstyle{rm}\Url}\fi

\bibitem[Abe and Warmuth(1992)]{DBLP:journals/ml/AbeW92}
N.~Abe and M.K. Warmuth.
\newblock On the computational complexity of approximating distributions by
  probabilistic automata.
\newblock \emph{Machine Learning}, 9:\penalty0 205--260, 1992.

\bibitem[Anandkumar et~al.(2012)Anandkumar, Hsu, and
  Kakade]{DBLP:conf/colt/Anand12}
A.~Anandkumar, D.~Hsu, and S.M. Kakade.
\newblock A method of moments for mixture models and hidden markov models.
\newblock In \emph{COLT}, 2012.

\bibitem[Baum et~al.(1970)Baum, Petrie, Soules, and Weiss]{baumwelch1970}
L.E. Baum, T.~Petrie, G.~Soules, and N.~Weiss.
\newblock A maximization technique occurring in the statistical analysis of
  probabilistic functions of {M}arkov chains.
\newblock \emph{Ann. Math. Stat.}, 41\penalty0 (1):\penalty0 pp. 164--171,
  1970.

\bibitem[Belkin and Sinha(2010)]{belkin2010polynomial}
M.~Belkin and K.~Sinha.
\newblock Polynomial learning of distribution families.
\newblock In \emph{Foundations of Computer Science (FOCS)}, pages 103--112,
  2010.

\bibitem[Bickel et~al.(1998)Bickel, Ritov, and Ryd{\'e}n]{MR1647705}
P.J. Bickel, Y.~Ritov, and T.~Ryd{\'e}n.
\newblock Asymptotic normality of the maximum-likelihood estimator for general
  hidden {M}arkov models.
\newblock \emph{Ann. Statist.}, 26\penalty0 (4):\penalty0 1614--1635, 1998.

\bibitem[Capp{\'e} et~al.(2005)Capp{\'e}, Moulines, and Ryd{\'e}n]{MR2159833}
O.~Capp{\'e}, E.~Moulines, and T.~Ryd{\'e}n.
\newblock \emph{Inference in hidden {M}arkov models}.
\newblock Springer Series in Statistics. Springer, New York, 2005.

\bibitem[Chang(1996)]{MR1410044}
J.T. Chang.
\newblock Full reconstruction of {M}arkov models on evolutionary trees:
  identifiability and consistency.
\newblock \emph{Math. Biosci.}, 137\penalty0 (1):\penalty0 51--73, 1996.

\bibitem[Cybenko and Crespi(2011)]{5773017}
G.~Cybenko and V.~Crespi.
\newblock Learning hidden {M}arkov models using nonnegative matrix
  factorization.
\newblock \emph{IEEE Trans. Information Theory}, 57\penalty0 (6):\penalty0 3963
  --3970, 2011.

\bibitem[Daniel(1973)]{springerlink:10.1007/BF01580110}
J.W. Daniel.
\newblock Stability of the solution of definite quadratic programs.
\newblock \emph{Mathematical Programming}, 5:\penalty0 41--53, 1973.

\bibitem[Dantzig et~al.(1967)Dantzig, Folkman, and Shapiro]{MR0207426}
G.B. Dantzig, J.~Folkman, and N.~Shapiro.
\newblock On the continuity of the minimum sets of a continuous function.
\newblock \emph{J. Math. Anal. Appl.}, 17:\penalty0 519--548, 1967.

\bibitem[den Hertog(1994)]{MR1282714}
D.~den Hertog.
\newblock \emph{Interior point approach to linear, quadratic and convex
  programming}, volume 277 of \emph{Mathematics and its Applications}.
\newblock Kluwer, Dordrecht, 1994.

\bibitem[Douc and Matias(2001)]{douc2001}
R.~Douc and C.~Matias.
\newblock Asymptotics of the maximum likelihood estimator for general hidden
  {M}arkov models.
\newblock \emph{Bernoulli}, 7\penalty0 (3):\penalty0 pp. 381--420, 2001.

\bibitem[Farag{\'o} and Lugosi(1989)]{MR1029474}
A.~Farag{\'o} and G.~Lugosi.
\newblock An algorithm to find the global optimum of left-to-right hidden
  {M}arkov model parameters.
\newblock \emph{Problems Control Inform. Theory/Problemy Upravlen. Teor.
  Inform.}, 18\penalty0 (6):\penalty0 435--444, 1989.

\bibitem[Hsu et~al.(2009)Hsu, Kakade, and Zhang]{DBLP:conf/colt/HsuKZ09}
D.~Hsu, S.M. Kakade, and T.~Zhang.
\newblock A spectral algorithm for learning hidden markov models.
\newblock In \emph{COLT}, 2009.

\bibitem[Kontorovich and Weiss(2012)]{meroi2012}
A.~Kontorovich and R.~Weiss.
\newblock Uniform {C}hernoff and {D}voretzky-{K}iefer-{W}olfowitz-type
  inequalities for {M}arkov chains and related processes, arxiv:1207.4678.
\newblock 2012.

\bibitem[Lakshminarayanan and Raich(2010)]{5589231}
B.~Lakshminarayanan and R.~Raich.
\newblock Non-negative matrix factorization for parameter estimation in hidden
  markov models.
\newblock In \emph{Machine Learning for Signal Processing (MLSP)}, pages 89
  --94, 2010.

\bibitem[Lyngs{\o} and Pedersen(2001)]{Lyngso:2009:CCH:1575125.1575172}
R.~B. Lyngs{\o} and C.~N. Pedersen.
\newblock Complexity of comparing hidden markov models.
\newblock In \emph{Proceedings of the 12th International Symposium on
  Algorithms and Computation}, pages 416--428. Springer-Verlag, 2001.

\bibitem[Mohri and Rostamizadeh(2010)]{mohri2010stability}
M.~Mohri and A.~Rostamizadeh.
\newblock Stability bounds for stationary $\varphi$-mixing and $\beta$-mixing
  processes.
\newblock \emph{The Journal of Machine Learning Research}, 11:\penalty0
  789--814, 2010.

\bibitem[Moitra and Valiant(2010)]{moitra2010settling}
Ankur Moitra and Gregory Valiant.
\newblock Settling the polynomial learnability of mixtures of gaussians.
\newblock In \emph{2010 IEEE 51st Annual Symposium on Foundations of Computer
  Science}, pages 93--102. IEEE, 2010.

\bibitem[Mossel and Roch(2006)]{MR2244426}
E.~Mossel and S.~Roch.
\newblock Learning nonsingular phylogenies and hidden {M}arkov models.
\newblock \emph{Ann. Appl. Probab.}, 16\penalty0 (2):\penalty0 583--614, 2006.

\bibitem[Nesterov and Nemirovskii(1994)]{MR1258086}
Y.~Nesterov and A.~Nemirovskii.
\newblock \emph{Interior-point polynomial algorithms in convex programming}.
\newblock SIAM, Philadelphia, PA, 1994.

\bibitem[Rabiner(1990)]{Rabiner:1990:THM:108235.108253}
L.~R. Rabiner.
\newblock Readings in speech recognition.
\newblock chapter A tutorial on hidden Markov models and selected applications
  in speech recognition, pages 267--296. Morgan Kaufmann, 1990.

\bibitem[Roweis and Ghahramani(1999)]{Roweis:1999:URL:309394.309396}
S.~Roweis and Z.~Ghahramani.
\newblock A unifying review of linear gaussian models.
\newblock \emph{Neural Comput.}, 11:\penalty0 305--345, February 1999.
\newblock ISSN 0899-7667.

\bibitem[Siddiqi et~al.(2010)Siddiqi, Boots, and Gordon]{Siddiqi10}
S.~M. Siddiqi, B.~Boots, and G.~J. Gordon.
\newblock Reduced-rank {H}idden {M}arkov {M}odels.
\newblock In \emph{AISTAT}, 2010.

\bibitem[Terwijn(2002)]{Terwijn:2002:LHM:645519.655953}
S.~Terwijn.
\newblock On the learnability of {H}idden {M}arkov {M}odels.
\newblock In \emph{Proceedings of the 6th International Colloquium on
  Grammatical Inference: Algorithms and Applications}, ICGI '02, pages
  261--268, London, UK, 2002. Springer-Verlag.

\end{thebibliography}
\bibliographystyle{plainnat}
\newpage
\section{Appendix}
We now give a detailed account for the theorems stated in section \ref{sec:stats}.

\subsection{Preliminaries I}
In what follows we use the following notation: For an $n\times n$ matrix
$A$,  $\vec(A)\in\R^{n^2}$ is the result of stacking
its columns vertically into a single long vector. 
Thus, its Frobenius matrix norm
is $\nrm{A}_F=\nrm{\vec(A)}_2$. 

Recall the definition of $g_\mix$:
\[
g_\mix \equiv \frac{ 2 G }{1-\mix}.
\]
One can easily verify that for $2G \geq 1,$ we have $1+\mix g_\mix \leq g_\mix^2$.
Also recall that assumption (2b) states that the distributions in $\F_n^\prm$ are bounded by $L$, which is defined by:
$$ \max_{i\in [n]} \sup_{y\in\R} f_{\prm_i}(y)\le L<\infty.$$

The following concentration result from \citet[Theorem 1]{meroi2012}
is our main tool in proving the error bounds given here.
\begin{lemma}
\label{lem:hmm-conc}
Let $Y=Y_0,\ldots,Y_{T-1}\in\Y^T$ be the output of a Hidden Markov chain with transition matrix $A$ and output distributions $\F_n^\prm$. Assume that $A$ is geometrically ergodic with constants $G,\mix$ as in (\ref{eq:geo-erg}).
Let $F:(Y_0,\ldots,Y_{T-1})\mapsto \R$
be any function that is $l$-Lipschitz with respect to the Hamming metric
on $\Y^T$.
Then, for all
$\eps>0$,
\beqn
\label{eq:kr}
P(|F(Y)-\E F|>\eps T) \le 2 \exp\paren{-\frac{T(1-\mix)^2\eps^2}{2l^2G^2}}.
\eeqn
\end{lemma}

We will also need the following Lemma (proved in \citep{meroi2012} for the discrete output case but easily generalize to continuous outputs) for bounding the variance of our estimators.

\begin{lemma}{\quad}
\label{lem:varBound}
Let $f(y):\R \to \R^+$ be a function of the observables of an $n$ states geometrically ergodic HMM with constants $(G,\mix)$ and 
\[
\int_\Y f(y) dy \leq 1.
\]
Assume the HMM is started with the stationary distribution $\bpi$.
 Then
\beq
\var\left[\oo{T}\sum_{t=0}^{T-1} f(Y_t)\right] 
& \leq & 
 \frac{\var[f(Y)]}{T} + \frac{ \mix g_\mix  \E[f(Y)]}{T}.
\eeq
Similarly, let $g(y,y'):\R \times \R \to \R^+$ be a function of {\bf consecutive} observations $(y,y')$ such that
\[
\iint_{\Y} g(y,y') dydy' \leq 1.
\]
 Then
\beq
\var\left[\oo{T}\sum_{t=1}^{T-1} g(Y_t,Y_{t+1})\right] 
& \leq & 
\frac{\var[g(Y,Y')]}{T-1} + \\
 &  &  \frac{ (1 + \mix g_\mix)  \E[g(Y,Y')]}{T-1}.
\eeq
\end{lemma}

\subsection{Accuracy of $\hat\bro,\hat\bsi,\hat\bxi$ and $\hat\bet$}
Since our estimators $\hat\bpi$ and $\hat A$ are constructed in terms of
$\hat\bro$ and $\hat\bsi$ in the discrete case, and $\hat\bxi$ and $\hat\bet$ in the continuous case, let us first examine the accuracy of the later. The following results
shows that 
geometric ergodicity
is sufficient to ensure their rapid
convergence to the true values.

\begin{lemma}
\label{lem:concDis}
{\bf Discrete case}. Let \((y_{t})_{t=1}^T\) be an observed sequence from a discrete output HMM whose initial state \(X_{0}\) 
follows the stationary distribution \(\bpi\).  
Let $\bro$ be given by (\ref{eq:rhodef}) and $\bsi$ by (\ref{eq:sigdef})
with their empirical estimates given in (\ref{eq:rhohat}).
Then
\beqn
\label{eq:E|rho|}
\E[ \nrm{\hat\bro-\bro}_2 ] & \le & \sqrt{\frac{1 + \mix g_\mix}{T}} \\
\label{eq:E|sig|}
\E[ \nrm{\hat\bsi-\bsi}_2 ] & \le & \sqrt{\frac{2 + \mix g_\mix}{T-1}}
\eeqn
Furthermore, for any $\eps > 0$ ,
\beqn
\label{eq:P|rho|}
&P(\nrm{\hat\bro-\bro}_2 > 
\sqrt{\frac{1 + \mix g_\mix}{T}}
 + \eps)
  \le  2 \exp\left({-\frac{2T\eps^2}{g_\mix^2}}\right) \quad
\eeqn
and
\begin{align}
\label{eq:P|sig|}
P  \left(
 \nrm{\hat\bsi-\bsi}_2
 > \sqrt{\frac{2 + \mix g_\mix}{T-1}} + \eps \right)& \le \\
\nonumber
   2 \exp & \left(\frac{-2(T-1)\eps^2}{g_\mix^2}\right).
\end{align}
Finally, we have for any fixed $\bv\in\R^m$ 
with $\nrm{\bv}_2=1$,
\beqn
P(\abs{\iprod{\hat\bro}{\bv}-\iprod{\bro}{\bv}}>\eps) 
\le 2 \exp\left(-\frac{2T\eps^2}{g_\mix^2}\right). 
        \label{eq:rho_v}
\eeqn
\end{lemma}

\bepf
First note that w.r.t the Hamming metric, $T||\hat{\bro}-\bro||_2$ and $\abs{\iprod{\hat\bro}{\bv}-\iprod{\bro}{\bv}}$ are 1-Lipschitz and $T||\hat{\bsi}-\bsi||_2$ is 2-Lipschitz. Thus the claims in (\ref{eq:P|rho|}, \ref{eq:P|sig|}, \ref{eq:rho_v}) all follows directly from Lemma \ref{lem:hmm-conc} where for (\ref{eq:P|rho|}, \ref{eq:P|sig|}) we also take into account (\ref{eq:E|rho|}) and (\ref{eq:E|sig|}) respectively.
 In order to prove (\ref{eq:E|rho|}) note that
\beq
\E[ \nrm{\hat\bro-\bro}_2^2] = \sum_{k\in[n]} \E(\hat{\rho}_k - \rho_k)^2 = \sum_{k\in[n]}Var(\hat{\rho_k}).
\eeq
So by taking  in Lemma \ref{lem:varBound}, $f(y) = \chr_{y = k}$, we have $\E[\chr_{y = k}] = \rho_k$  and $Var(\chr_{y = k}) ={\rho_k (1-\rho_k)} \leq \rho_k$. Since $\sum_{k=1}^m \rho_k = 1$ we get the desired bound.

The bound in (\ref{eq:E|sig|}) is obtained similarly by taking $g(y,y') = \chr_{y = k}\chr_{y' = k'}$ in Lemma \ref{lem:varBound} with the fact that $\sum_{kk'} \sigma_{kk'} = 1$.
\enpf

\begin{lemma}
\label{lem:concPar}
{\bf Continuous case}. 
Let \((Y_t)_{t=1}^T\) be an observed sequence from a continuous observations HMM whose initial state \(X_{0}\) 
follows the stationary distribution \(\bpi\).  
Let $\bxi$ be given by (\ref{eq:xi}) , $\bet$ by (\ref{eq:Etakk}) and
$\hat\bxi$ and $\hat\bet$ be their empirical estimates, given by (\ref{eq:xihat}) and (\ref{eq:etahat}) respectively.
Then
\hide{
\beqn
\label{eq:E|xi|}
\E[ \nrm{\hat\bxi-\bxi}_2 ] & \le & \sqrt{\frac{n L(L + g_\mix)}{T}},\\
\label{eq:E|eta|}
\E[ \nrm{\hat\bet-\bet}_2 ] & \le & \sqrt{\frac{g_\mix}{T-1}}.
\eeqn
Furthermore,
}
for any $\eps > 0$ ,
\beqn
\label{eq:P|xi|}
P\left(\nrm{\hat\bxi-\bxi}_2>
 \eps \right) \le 
 2 n \exp\left({-\frac{2T\eps^2}{g_\mix^2 nL^2}}\right),
\eeqn
and
\begin{align}
\label{eq:P|eta|}
P  \left(
 \nrm{\hat\bet-\bet}_2
 > \eps \right)& \le \\
\nonumber
   2 n^2\exp & \left(-\frac{2(T-1)\eps^2}{g_\mix^2n^2}\right).
\end{align}
\end{lemma}

\bepf
Note that $\E \hat{\xi}_k = \xi_k$ and $T\hat{\xi}_k$ is $L$-Lipschitz for all $k \in [n]$. Thus by Lemma \ref{lem:hmm-conc} and the union bound we have
\beqn
\label{eq:P_infty_xi}
P\left(\nrm{\hat\bxi-\bxi}_\infty>
 \eps' \right) \le 
 2 n \exp\left({-\frac{2T\eps'^2}{g_\mix^2 L^2}}\right).
\eeqn
Since
\beq
\nrm{\hat\bxi-\bxi}_2^2 = {\sum_{k\in[n]} (\hat{\xi}_k - \xi_k)^2} 
\leq n \nrm{\hat\bxi-\bxi}_\infty^2,
\eeq
we have
\beq
P\left(\nrm{\hat\bxi-\bxi}_2>
 \eps \right)
 \le
 P\left(\sqrt n\nrm{\hat\bxi-\bxi}_\infty>
 \eps \right).
\eeq
putting $\eps' = \eps /\sqrt n$ in (\ref{eq:P_infty_xi}), the claim in (\ref{eq:P|xi|}) follows.

The proof of (\ref{eq:P|eta|}) follows the same 
paradigm as the proof for (\ref{eq:P_infty_xi}). Indeed $\E[\hat{\eta}_{kk'}] = \eta_{kk'}$ and $T\hat{\eta_{kk'}}$ is $1$-Lipschitz so by Lemma \ref{lem:hmm-conc} and the union bound we have
\beqn
\label{eq:P_infty_eta}
P\left(\nrm{\hat\bet-\bet}_\infty>
 \eps' \right) \le 
 2 n^2 \exp\left({-\frac{2T\eps'^2}{g_\mix^2 L^2}}\right).
\eeqn
Since
\beq
\nrm{\hat\bet-\bet}_2^2 = {\sum_{k,k'\in [n] \times [n]} (\hat{\eta}_{kk'} - \eta_{kk'})^2} 
\leq n^2 \nrm{\hat\bet-\bet}_\infty^2,
\eeq
we have
\beq
P\left(\nrm{\hat\bet-\bet}_2>
 \eps \right)
 \le
 P\left( n\nrm{\hat\bet-\bet}_\infty>
 \eps \right).
\eeq
putting $\eps' = \eps / n$ in (\ref{eq:P_infty_eta}), the claim in (\ref{eq:P|eta|}) follows.
\enpf

\subsection{Proof of theorem \ref{thm:consist} - Strong consistency }
We now prove the strong consistency of our estimators stated in Theorem \ref{thm:consist}.
\bepf
For the discrete case, 
by Lemma \ref{lem:concDis}, the expectation $\E[ \nrm{\hat\bro-\bro}_2 ]$ goes to zero as $T \rightarrow \infty$. Furthermore, using the Borel-Cantelli lemma, $\nrm{\hat\bro-\bro}_2$ converge to its expectation a.s. concluding that  $\hat\bro$ converges a.s. to $\bro$. The same argument goes for $\hat\bsi$, $\hat\bxi$, $\hat\bet$ and   $\bsi$, $\bxi$, $\bet$ respectively.

Now, the function $f:\R^m\to\R^n$ given by $f(x)=(B\trn\diag(1/x)B)\inv\boldsymbol{1}$
is continuous on $\R_+^m$.
Moreover,
$f(\bro)=\bpi$ since the optimization problem
(\ref{eq:QP_x}) has a unique minimizer $x^*$ for all $\hat\bro$, which 
in particular is given by $x^*=\bpi$ when $\hat\bro=\bro$.
Since 
$\bro\in\R_+^m$
by 
assumption,
the argument above shows
that almost surely, $\hat\bro\in\R_+^m$ for all sufficiently large $T$. Therefore, 
$ \lim_{T\to\infty} f(\hat\bro) = f(\bro)= \bpi$
almost surely, and the asymptotic strong consistency of $\hat\bpi$ is established.

To prove the
asymptotic strong consistency of $\hat A$ 
in the discrete case,
recall that the minimizer of the 
quadratic program
$x\trn Kx-h\trn x$
subject to $Gx\le g$, $Dx=d$,
is continuous under small perturbations of $K,h,G,D,d$ \citep{MR0207426}.
In particular,  if $\hat\bpi$ 
is sufficiently
close to $\bpi$
then
$\hat A$
is close to
$A$.
Since $\hat\bpi\to\bpi$ and  $\hat\bsi\to\bsi$ almost surely, we also have 
$
\hat A
{\overset{\textrm{\tiny \textup{a.s.}}}{\longrightarrow}}
A
$.

For the continuous observations case, note that $\hat \pi$ and $\hat A$ are also solutions of quadratic programs. Also note that $\hat\bxi\to\bxi$ and  $\hat\bet\to\bet$ almost surely. Thus we have that 
$
\hat A
{\overset{\textrm{\tiny \textup{a.s.}}}{\longrightarrow}}
A
$
and
$
\hat \bpi
{\overset{\textrm{\tiny \textup{a.s.}}}{\longrightarrow}}
\bpi
$
as above.
\enpf

\subsection{{Proof of Theorem \ref{thm:|pi|}: Bounding the error for $\hat\pi$} in the discrete observations case}
\bepf
Lemma \ref{lem:concDis} and the fact that \(\|\hat\bro-\bro\|_\infty \leq \|\hat\bro-\bro\|_2\) implies
that \(\|\hat\bro-\bro\|_\infty=O_P(1/\sqrt{T}).\) 
Hence we make a change of variables, 
\begin{equation}
\hat\bro=\bro+\frac1{\sqrt{T}}\zeta.
\end{equation}
To establish the (eventual) positivity of the entries of $\hat\bpi$, 
we consider the solution \(x^*\) of 
(\ref{eq:QPpi}) with \(\lambda=0\), e.g. without the normalization $\sum x_i=1$, and write it as
$x^*=\bpi+\delta$. 
Our goal is to understand the relation between $\delta$ and $\zeta$.  

Observe that $\delta$ satisfies 
the system of linear equations
\beq
\sum_j\Big(
 \sum_k \frac{B_{kj}B_{ki}}
{\rho_k\paren{1+\frac1{\sqrt{T}}\frac{\zeta_k}{\rho_k} }}\Big)
(\pi_j+\delta_j)=1.
\eeq
We need $T$ sufficiently large so that, with high probability, 
$\max_k\frac1{\sqrt{T}}\frac{\zeta_k}{\rho_k}\ll 1$,
or equivalently,
$\abs{\hat\rho_k-\rho_k}\ll\rho_k$. 

By taking \(T
\gtrsim 
4g_\mix / a_1^2 \) we have 
\beq
\E[ \nrm{\hat\bro-\bro}_\infty ] \leq a_1/2.
\eeq
 
So choosing $\eps=\min\rho_k /2\ge a_1/2$ in (\ref{eq:P|rho|}), this condition
is satisfied for 
\(T
\gtrsim 
g_\mix^2 / a_1^2 \). 
Then, approximating $1/(1+\epsilon)=1-\epsilon+O(\epsilon^2)$ gives 
\beq
 \sum_j\sqprn{
 \sum_k \frac{B_{kj}B_{ki}}{\rho_k}
\paren{1-\frac{1}{\sqrt{T}}\frac{\zeta_k}{\rho_k}}}(\pi_j+\delta_j)\\
  =  1+O_P\paren{\oo T}.
\eeq
Note that since \(B\bpi=\bro\), the leading order correction for $\delta$ is simply 
\beq
\delta = \oo{\sqrt T}(\tilde B\trn\tilde B)\inv\tilde B\trn
\left(\frac{\zeta}{\bro}\right)
+O_P\paren{\oo T},
\eeq
where the matrix $\tilde B = \diag(1/\sqrt{\bro})B$. 
 
Let $\set{\bu_{i}}$ and $\set{\bv_i}$ be the right and left singular vectors of $\tilde B$ with non-zero singular values $\sigma_i(\tilde B)$, where $\sigma_1\leq \sigma_2\ldots \leq \sigma_n$; 
thus, $\tilde B\bu_i = \sigma_i \bv_i$. 
The fact that $\tilde B$ also has $n$ non-zero singular values follows from its definition combined with 
our
Assumption 2d that $B$ has rank $n$.
Then
\begin{equation}
\tilde B^{\trn}\tilde B = \sum_i \sigma_i^2 \bu_i \bu_i^{\trn}
\end{equation}
and hence,
\begin{equation}
\delta =\frac1{\sqrt{T}} \sum_i \frac{1}{\sigma_i}\langle{\frac{\zeta}{\bro}},{\bv_i}\rangle \bu_i +O_P\left(\frac1T\right)
\end{equation}
For the solution \(x\) to have strictly positive coordinates we need 
that $\abs{\delta_j} < \pi_j$
for each of $j=1,\ldots,n$.
Without loss of generality, assume that $\pi_1=\min_j \pi_j$ and analyze the worst-case setting. 
This occurs when the singular vector \(\bu_1\) with smallest singular value coincides with the standard basis vector ${\bf e}_1$. 
Then, 
\begin{equation}
|\delta_1| \leq \frac{1}{\sqrt{T}}\frac1{\sigma_1(\tilde B) \min_j \rho_j}|\langle \zeta,\bv_1\rangle|+O_{P}\left(\frac1T\right) .
        \label{eq:abs_delta1}
\end{equation}
It follows from 
(\ref{eq:rho_v}) that \(|\delta_1|\) will be dominated by
\(\min \pi_j \geq a_0 \)
provided
that
\beqn
T\gtrsim
\frac{g_\mix}{ a_0 a_1 \sigma_1(\tilde B)}.
\eeqn
In the unlikely event
that 
(i)
the vector \(\bpi\) is uniform ($\pi_j=1/n$ for all $j$),  
(ii) the matrix $\tilde B$ has $n$ identical singular values, 
we need the equation analogous to (\ref{eq:abs_delta1}) to hold for all $n$ coordinates. 
By a union bound argument, 
an additional factor of $\log n$ in the number of samples suffices to 
ensure, with high probability,
the non-negativity 
of the solution \(x\). 

Next we proceed to bound $\nrm{\hat\bpi-\bpi}_2^2$. To this end, we write
\beq
x^*-\bpi = \delta = \sum_{i}\oo{\sigma_i(\tilde B)}
\langle\frac{\hat\bro-\bro}{\bro},\bv_i\rangle 
\bu_i + O_P\paren{\oo T}.
\eeq
Since both the $\set{\bu_i}$ and the $\set{\bv_i}$ are orthonormal,
\beq
\nrm{\delta}_2^2 &=&
\sum_i \oo{\sigma_i^2(\tilde B)}%
\langle\frac{\hat\bro-\bro}{\bro},\bv_i\rangle 
^2\\
&\le &
\oo{\sigma_1^2(\tilde B)(\min\rho_k)^2}
\sum_i
\langle\hat\bro-\bro,\bv_i\rangle %
^2\\
& \le &
\frac{\nrm{\hat\bro-\bro}_2^2}{\sigma_1^2(\tilde B)a_1^2}.
\eeq
\hide{
By Lemma \ref{lem:conc},  \(\forall i\in[m]\),
$ \E(\hat\rho_i-\rho_i)^2 \le \frac{G^2}{T(1-\mix)^2}$.
}
Bounding $\nrm{\hat\bro-\bro}_2^2$ via Lemma \ref{lem:concDis} and noting that
$$
\nrm{\hat\bpi-\bpi}_2 =
\nrm{\tfrac{x^*}{\nrm{x^*}_1}
-\bpi}_2
\leq 2 \nrm{x^*-\bpi}_2=2\nrm{\delta}_2,
$$
 the result in
(\ref{eq:Error_pi}) follows. 
\enpf

\subsection{Preliminaries II}
The remaining estimators ($\hat{\pi}$ for the continuous observations case, and $\hat{A}$ for both the discrete and continuous observations cases) are obtained as solutions for quadratic programs.
Let us take for example the QP for calculating $\hat\bpi$ with continuous observations HMM, given in (\ref{eq:QPpiParam}). For this case, the QP is equivalent to
\beq
\hat{\bpi} = \arg\!\min_x  \frac12 x^{\trn} {K}^{\trn} {K} x - x^{\trn}{K}^{\trn} \hat{\bxi}
\eeq
subject to $x \geq 0$ and $\sum_i x_i = 1$.

Note that if \(\hat\bxi\) was equal to its true values \(\bxi\), the solution of the above QP would simply be the true $\bpi$. In reality, we only have the estimate $\hat{\bxi}$.
In order to analyze the error $\nrm{\hat{\bpi} - \bpi}_2$, we will need to consider how the solutions of such a quadratic program are affected by errors in  $\bxi$. 

More generally, we are concerned with two QPs
\beqn
\label{eq:QP_x_appendix}
   \min {Q}(x) & = &  \min\frac12 x^{\trn} {M} x-x^{\trn} {h},\\
\label{eq:QP_hat_x_appendix}
   \min \hat{Q}(x) & = &  \min\frac12 x^{\trn} \hat{M} x-x^{\trn}\hat{h},
\eeqn
both subject to $Gx  \le  g$, $Dx  =  d$. We assume that the solution to the first QP is the ``true'' value while the solution to the second is our estimate. So bounding the estimate error is equivalent to bounding the error between the solutions obtained by the above two QPs, where $\hat{M}$ and $\hat{h}$ are perturbed versions of $M$ and $h$.

Given that, note that only the objective function has been perturbed, while the linear constraints remained unaffected.
We may thus apply the following classical result on the solution stability of definite quadratic programs.

\begin{theorem} \citep{springerlink:10.1007/BF01580110} 
\label{thm:daniel}
Let $\lambda=\lambda_{\min}(M)$ be the smallest eigenvalue of $M$, and let 
$\epsilon=\max\{\|\hat M-M\|_2,\|\hat h-h\|_2\}$. Let $x$ and $\hat{x}$ 
be the minimizers of Eqs.(\ref{eq:QP_x_appendix}) and (\ref{eq:QP_hat_x_appendix}), 
respectively. Then, for $\epsilon<\lambda$,
\beq
\| x-\hat x\|_2 \leq \frac{\epsilon}{\lambda - \epsilon}
(1+\|x\|_2).
\eeq
\end{theorem}

In the following we will obtain bounds on $\eps$ and $\lambda$  for the different estimators and invoke the above theorem. 
\subsection{{Proof of Theorem \ref{thm:|pi|Par}: Bounding the error for $\hat\pi$} in the continuous observations case}

\bepf
Note that in the notation given in Theorem \ref{thm:daniel}, we have
\(h =  \bxi^{\trn}K\) and \(\hat{h} = \hat{\bxi}^{\trn} {K}\). 
Since we assumed that the output density parameters are known exactly we have no error in $M = K^{\trn} K$.

It is immediate that 
\[
\lambda_{min}(K^{\trn} K) = \sigma_1^2(K),
\]
and
\[
\eps \leq \nrm{\hat{\bxi} - \bxi}_2 \nrm{K}_2 \leq n L \nrm{\hat{\bxi} - \bxi}_2.
\]

From Lemma \ref{lem:concPar} we have
\[
\nrm{\hat{\bxi} - \bxi}_2 \lesssim_P \sqrt{\frac{(n \ln n) g_\mix^2 L^2}{T }},
\]

while by Theorem \ref{thm:daniel} we have
\[
 \nrm{\hat\bpi-\bpi}_2 \lesssim \frac{\eps}{\lambda_{min}(K^{\trn} K)}(1 + \nrm{\bpi}_2).
\]
Since $\nrm{\bpi}_2 \leq 1$, the claim follows.
\enpf

As a side remark we note that the form of (\ref{eq:Error_pi_param}) is somewhat counter-intuitive, as
it suggests a worse behavior for larger $L$. Intuitively, however, larger $L$ corresponds to a more peaked --- and hence lower-variance ---
density, which ought to imply sharper estimates.
Note however that as numerical simulations suggest we typically have
\beq
\frac{\sigma_1^2(\tilde F) L^2}{\sigma_1^2(\tilde K)}  = O(1).
\eeq
Thus, whenever $\sigma_1^2(\tilde F)$ is well behaved so is the estimate in 
(\ref{eq:Error_pi_param}) and the bound is reasonable after all. Finally note that $F$ is stochastic so it behaves very much like the matrix $B$ in the discrete outputs case.

\subsection{Proof of Theorem \ref{thm:main}: Bounding the error of $\hat A$ in the discrete observations case}
Let $\hat A$ be the solution of
\begin{equation*}
\min_{A_{ij}\geq 0,\sum_i A_{ij}=1} \|\hat\bsi-\hat C A \|^2_2, \tag{\ref{eq:|CA|}} 
\end{equation*}
where $\hat{\bsi}$ is given in (\ref{eq:rhohat}). Recall that ${C}_{ij}^{kk'}={\pi}_j B_{kj}B_{k'i}$ and $\hat{C}_{ij}^{kk'}=\hat{\pi}_j B_{kj}B_{k'i}$.
First note that if
\(\bpi\) and \(\bsi\) were known exactly, the above QP could be written as 
\beqn
\min Q(A)=\min\frac12 \vec(A)^{\trn} M \vec(A)-\vec(A)^{\trn}h
        \label{eq:QP_A}
\eeqn
where $M=C^{\trn}
C$ and $h=C^{\trn}
\vec(\bsi)
$. 
Its solution is 
precisely
the transition probability matrix $A$. 
In reality, as we only have estimates $\hat\bpi$ and $\hat\bsi$, 
the optimization problem is perturbed to 
\beqn
\min \hat Q(A)=
\min\frac12 \vec(A)\trn\hat{M}\vec(A)-\vec(A)^{\trn}\hat h
        \label{eq:QP_hat_A}
\eeqn
where $\hat M = \hat C^{\trn}\hat C$, 
and $\hat h = \hat C^{\trn}\vec(\hat\bsi)$.

To analyze how errors in  $\hat\bsi$ and $\hat C$ affect the optimization
problem we follow the same route as above. Thus we need to bound
$\|\hat h-h\|_2$, $\|\hat M - M\|_2$, and the smallest eigenvalue of $M$. 
Regarding the latter, by definition, 
$\lambda_{\min}(M)=\sigma_1^2( C)$, where $\sigma_1(C)$ is the smallest singular value of ${C}$. 
A simple exercise in linear algebra yields
\beqn
\label{eq:sig(C)} 
\sigma_1({C}) \geq 
a_0\sigma_1^2({B})
.
\eeqn
The following lemma provides bounds on $\|\hat M - M\|_2$ and on $\|\hat h - h\|_2$.
\begin{lemma} 
\label{lem:|Kb|}
Asymptotically, as $T\to\infty$,
\beqn
\|\hat h - h \|_{2} 
\lesssim_P
\sqrt{n}\left(
\|\hat\bpi-\bpi\|_2+ 
\nrm{\hat\bsi-\bsi}_2
\right) 
\label{eq:ERROR_b}
\eeqn
and
\beqn
\|\hat M - M\|_2 
\lesssim_P
2 n \|\hat\bpi - \bpi\|_2 
.
\label{eq:ERROR_K}
\eeqn
\end{lemma}

\bepf
By definition, $h_{ij}=\sum_{k,k'}C^{kk'}_{ij}\sigma_{kk'}$, and 
$\hat h_{ij}=\sum_{k,k'}\hat C^{kk'}_{ij}\hat\sigma_{kk'}$. 
Using the definitions of $C$ and $\hat C$, up to mixed 
terms \(O(\|\hat\bpi-\bpi\|_\infty\|\hat\bsi-\bsi\|_\infty)\), we obtain
\beq
\hat h_{ij}-h_{ij} &=& (\hat \pi_j-\pi_j)\sum_{kk'}B_{kj}B_{k'i} \sigma_{kk'} \\
&& + \pi_j\sum_{kk'}B_{kj}B_{k'i} (\hat\sigma_{kk'}-\sigma_{kk'}) 
\eeq
Since each of $\nrm{\hat\bpi-\bpi}_\infty$ and $\nrm{\hat\bsi - \bsi}_\infty$ are $O_{P}(1/\sqrt{T})$, 
the neglected mixed terms are asymptotically negligible as compared to each of the first two ones. Next, we use the fact that 
$\sigma_{kk'}\leq 1, \pi_{j}\leq 1$ and $\sum_{kk'} B_{kj}B_{k'i}\leq 1$
to obtain that
\beqn
\nonumber
\nrm{\hat h - h}_2
\lesssim_P\sqrt{n}\nrm{\hat\bpi-\bpi}_2 +  \sqrt{n}\nrm{\vec(\hat\bsi) - \vec(\bsi)}_2
\eeqn
Similarly, we have that for the \(n^{2}\times n^2\) matrix $M$, and not including higher order mixed terms \((\hat\pi_j-\pi_j)(\hat\pi_\beta-\pi_\beta)\), which are asymptotically negligible, 
\beq
(\hat M - M)_{ij,\alpha\beta}= (\hat\pi_j - \pi_j)\pi_\beta
\sum_{kk'}B_{kj}B_{k\beta}
B_{k'i}B_{k'\alpha} 
\\
+(\hat\pi_\beta - \pi_\beta)\pi_j
\sum_{kk'}B_{kj}B_{k\beta}
B_{k'i}B_{k'\alpha} 
\eeq
Note that $\sum_{kk'}B_{kj}B_{k\beta}
B_{k'i}B_{k'\alpha} = (\sum_k B_{kj}B_{k\beta})( \sum_{k'}B_{k'i}B_{k'\alpha})
\leq 1$. Hence, by similar arguments as for $h$, (\ref{eq:ERROR_K}) follows.  
\enpf

We can now prove Theorem \ref{thm:main}:
\bepf ({\bf of Theorem \ref{thm:main}})
Lemma \ref{lem:concDis}, together with
(\ref{eq:Error_pi}),
implies that with high probability,
$$\nrm{\hat\bsi-\bsi}_F\lesssim_P \sqrt\frac{g_\mix^2}{T-1},
$$
and
$$
\nrm{\hat\bpi-\bpi}_2 \lesssim_P
\sqrt\frac{g_\mix^2}{T a_1^2 \sigma_1^2(\tilde B)}.
$$
Inserting these  into 
(\ref{eq:ERROR_b}) 
and
(\ref{eq:ERROR_K}) 
yields, w.h.p.,
\beqn
\nonumber
\label{eq:epsbd}
\eps & = &
\max\set{\nrm{\hat h-h}_2 
,
\nrm{\hat M-M}_2 }\\
& \lesssim &
\sqrt\frac{n^2 g_\mix^2}{T a_1^2 \sigma_1^2(\tilde B)}
.
\eeqn
By Theorem \ref{thm:daniel}, we have that 
\beqn
\label{eq:dan}
\nrm{\hat A-A}_F\lesssim \frac{\eps}{\lambda_1(M)}(1+\|A\|_F),
\eeqn
where $\|A\|_F\leq \sqrt{n}$ since $A$ is column-stochastic. 
The claim follows by substituting the bounds on $\eps$ in 
(\ref{eq:epsbd}) and on $\lambda_1(M)=\sigma_1^2(C) \geq a_0^2 \sigma_1^4(B)$ in (\ref{eq:sig(C)})
into (\ref{eq:dan}) and noting that $\sigma_1^2(\tilde B) \geq \sigma_1^2(B)$.
\enpf

\subsection{Proof of Theorem \ref{thm:mainPar}: Bounding the error of $\hat A$ in the continuous observations case}

Let $\hat A$ be the solution of
\begin{equation*}
\min_{A_{ij}\geq 0,\sum_i A_{ij}=1} \|\hat\bet-\hat C A \|^2_2, \tag{\ref{eq:|CA|}} 
\end{equation*}
where $\hat{\bet}$ is given in (\ref{eq:etahat}) and ${C}_{ij}^{kk'}={\pi}_j F_{kj}F_{k'i}$ and $\hat{C}_{ij}^{kk'}=\hat{\pi}_j F_{kj}F_{k'i}$.
The above QP can be written as
\beqn
\min \hat Q(A)=
\min\frac12 \vec(A)\trn\hat{M}\vec(A)-\vec(A)^{\trn}\hat h
        \label{eq:QP_hat_A_appendix}
\eeqn
where $\hat M = \hat C^{\trn}\hat C$, 
and $\hat h = \hat C^{\trn}\vec(\hat\bsi)$. 

Exactly as in the previous subsection, we want to bound the difference between the solutions for the above QP and the unperturbed one. 

First note that
\beqn
\label{eq:sig(C)Param} 
\sigma_1({C}) \geq 
a_0\sigma_1^2({F})
.
\eeqn
Next we give the analogue of lemma \ref{lem:|Kb|}.
\begin{lemma}
Asymptotically, as $T\to\infty$,
\beqn
\|\hat h - h \|_{2} 
\lesssim_P
\sqrt{n}\left(
\oo{a_0}
\|\hat\bpi-\bpi\|_2+ 
\nrm{\hat\bet-\bet}_2
\right) 
\label{eq:ERROR_bPar}
\eeqn
and
\beqn
\|\hat M - M\|_2 
\lesssim_P
2 n \frac{\|\hat\bpi - \bpi\|_2 }{a_0}
.
\label{eq:ERROR_KPar}
\eeqn
\end{lemma}

\bepf
In contrast to Lemma \ref{lem:|Kb|}, here $F$ is also perturbed due to errors in $\hat\bpi$ with
\[
\hat{F}_{ij} = \int_\Y \frac{\hat{\pi}_i f_i(y)f_j(y)}{\sum_k \hat{\pi}_k  f_k(y)} dy.
\]
Expending the difference $\Delta F_{ij} \equiv \abs{\hat{F}_{ij} - F_{ij}}$ up to first order in $\hat{\bpi} - \bpi$ we find that
\[
\nrm{\Delta F}_F \leq \frac{\nrm{\hat{\bpi} - \bpi}_\infty}{a_0} \nrm{F}_F 
\leq \frac{\sqrt n\nrm{\hat{\bpi} - \bpi}_\infty}{a_0},
\]
where in the last inequality we used the fact that $F$ is stochastic. Repeating the arguments in the proof for Lemma \ref{lem:|Kb|} and noting that $a_0 \ll 1$ we get (\ref{eq:ERROR_bPar}) and (\ref{eq:ERROR_KPar}).
\enpf

We now come to the proof of Theorem \ref{thm:mainPar}.
\bepf ({\bf of Theorem \ref{thm:mainPar}})
Lemma \ref{lem:concPar}, together with
(\ref{eq:Error_pi_param}),
implies that with high probability,
$$\nrm{\hat\bet-\bet}_F\lesssim_P \sqrt\frac{(n^2 \ln n) g_\mix^2}{T-1},
$$
and
$$
\nrm{\hat\bpi-\bpi}_2 \lesssim_P
\sqrt\frac{(n^3 \ln n)  g_\mix^2 L^4 }{ T \sigma_1^4(\tilde K) }
$$
Inserting these  into 
(\ref{eq:ERROR_bPar}) 
and
(\ref{eq:ERROR_KPar}) 
yields, w.h.p.,
\beqn
\nonumber
\label{eq:epsbdParam}
\eps & = &
\max\set{\nrm{\hat h-h}_2 
,
\nrm{\hat M-M}_2 }\\
& \lesssim &
\sqrt\frac{(n^5 \ln n) g_\mix^2 L^4}{T \sigma_1^4(\tilde K)}
.
\eeqn
By Theorem \ref{thm:daniel}, we have that 
\beqn
\label{eq:danParam}
\nrm{\hat A-A}_F\lesssim \frac{\eps}{\lambda_1(M)}(1+\|A\|_F),
\eeqn
where $\|A\|_F\leq \sqrt{n}$ since $A$ is column-stochastic. 
The claim follows by substituting the bounds on $\eps$ in 
(\ref{eq:epsbdParam}) and on $\lambda_1(M)=\sigma_1^2(C) \geq a_0^2 \sigma_1^4(F)$ in (\ref{eq:sig(C)})
into (\ref{eq:danParam}) and noting that $\sigma_1^2(\tilde F) \geq \sigma_1^2(F)$.
\enpf

As for remark \ref{rem:Kestimation}, we point out that estimating $\bet'$ with the help of the matrix $K$ (instead of $\bet$ with $F$) results in an estimator
that is not $O(1/T)$-Lipschitz any more but $O(L^2/T)$-Lipschitz with $L=\max_{i\in[n]}\sup_{y\in \R} f_{\prm_i}(y)$. This means that in principle we will need many more samples to accurately estimate $\bet'$ compared to $\bet$, see Lemma \ref{lem:concPar}.
Thus, since in high dimensions calculating $F$ via numerical integration may be computational intensive, choosing between the two estimators is in some sense choosing between working with limited number of samples and computational efficiency.

\subsection{Proof of Theorem \ref{thm:Bperturbed}:
Perturbations in the output parameters}
We give here the proof for the perturbation in the matrix $F$. The proof for perturbations in the matrix $K$ is similar.
\bepf
By definition,
$b_{ij} = \sum_{k,k'} C_{ij}^{kk'} \sigma_{kk'}$
, and
$\hat{b}_{ij} = \sum_{k,k'} \hat{C}_{ij}^{kk'} \hat{\sigma}_{kk'}$.
Using the definitions of $C$ and $\hat{C}$, up to first order in $\{\nrm{\hat{\pi} -\pi}_\infty,\nrm{\hat{\sigma} -\sigma}_\infty,\eb \}$ we obtain
\beq
\hat{b}_{ij} - b_{ij} =
(\hat{\pi}_j - \pi_j)\sum_{kk'}B_{kj}B_{k'i} \sigma_{kk'}
\\
+
\pi_j\sum_{kk'}B_{kj}B_{k'i}(\hat{\sigma}_{kk'} - \sigma_{kk'})
\\
+
\eb \pi_j
 \sum_{kk'}\left(
P_{kj}B_{k'i} + B_{kj}P_{k'i} 
\right)
\sigma_{kk'}.
\eeq

As the two first terms already considered we focus on the last term. It can be shown that:
\beq
\sum_{ij} 
\left(
\pi_j
\sum_{kk'} P_{kj}B_{k'i} \sigma_{kk'}
\right)^2
        & \leq &
n\nrm{P}_F^2.
\eeq
Thus
\beqn
\nrm{\hat{b} - b}_2 
\leq 
\sqrt{n} 
\left(
\nrm{\hat{\pi} - \pi}_2 + \nrm{vec(\hat{\sigma})- vec(\sigma)}_2 + \right. \\
\nonumber
 \left.  + 2\eb \nrm{P}_F
\right)
(1+o(1)).
\eeqn
Similarly, for the matrix $K$ up to first order in $\{\nrm{\hat{\pi} - \pi}_\infty ,\eb \}$ we have
\beq
(\hat{K} - K)_{ij,\a\b} 
& = &
(\hat{\pi}_j - \pi_j)\pi_\b
\sum_{kk'}
B_{kj}B_{k\b} B_{k'i}B_{k'\a}
\\
& + &
(\hat{\pi}_\b - \pi_\b)\pi_j
\sum_{kk'}
B_{kj}B_{k\b} B_{k'i}B_{k'\a}
\\
& + &
\eb \pi_j\pi_\b
\sum_{kk'}
P_{kj}B_{k\b} B_{k'i}B_{k'\a}
+ \dots 
\\
& + &
\eb \pi_\b\pi_j
\sum_{kk'}
B_{kj}B_{k\b} B_{k'i}P_{k'\a}.
\eeq

Again considering only the terms including $P$ and using the facts that $\sum_{k}B_{kj}B_{k\b} \leq 1$ and $\sum_{kk'} (P_{kj} B_{k'i})^2 \leq \sum_k P_{kj}^2$ we similarly find that
\beq
\nrm{\hat{K} - K}_2
\leq
(1+o_p(1))2n
\left(
\nrm{\hat{\pi} - \pi}_2 + 4 \eb \nrm{P}_F
\right).
\eeq

Repeating the analysis in the proofs for Theorems \ref{thm:|pi|Par}, \ref{thm:main} and \ref{thm:mainPar} give the desired result.
\enpf
\hide{
\subsection{Proof of Lemma \ref{lem:varBound}}
\label{sec:varLemma}
\bepf
By definition
\beq
& & \var \left[  \sum_{i=1}^T f(Y_i)\right] =  \\
& & \qquad   \E\left[\left(\sum_{i=1}^T f(Y_i)\right)^2\right] - \left(\sum_{i=1}^T \E [f(Y_i)]\right)^2\\
& & \qquad   \E\left[ \left(\sum_{i=1}^T f(Y_i)\right)^2\right] - T^2 \left( \E [f(Y)]\right)^2.
\eeq
Now
\beq
& &\E\left[ \left(\sum_{i=1}^T f(Y_i)\right)^2\right] = \\
& & \quad = \sum_{i=1}^T\E[f^2(Y_i)] + 2 \sum_{1 \leq i < j \leq T}\E[f(Y_i)f(Y_j)]\\
& & \quad =  T \E[f^2(Y)] + 2 \sum_{1 \leq i < j \leq T}\E[f(Y_i)f(Y_j)].
\eeq
In order to bound $\E[f(Y_i)f(Y_j)]$ it can be shown that
\beq
\nrm{P(Y_j = y \gn Y_i = y') - P(Y_j = y)}_\infty \leq G\theta^{j-i}.
\eeq

Indeed
\beq
& &\nrm{P(Y_j = y \gn Y_i = y') - P(Y_j = y)}_\infty \\
&  & \quad \leq
\nrm{P(Y_j = y \gn Y_i = y') - P(Y_j = y)}_{TV} \\
& & \quad = \oo{2} \int_\Y dy \left|P(Y_j = y \gn Y_i = y') - P(y)\right| \\
& & \quad = \oo{2} \int_\Y dy \left|\sum_{k\in[n]} P(y \gn k ) P(X_j = k \gn Y_i = y') - \sum_{k\in[n]} P(y \gn k) \pi_k \right| \\
& & \quad \leq \oo{2} \sum_{k\in[n]} \int_\Y dy P(y \gn k) 
\left| P(X_j = k \gn Y_i = y') - \pi_k \right|\\
& & \quad = \oo{2} \sum_{k\in[n]} \left| P(X_j = k \gn Y_i = y') - \pi_k \right|\\
& & \quad = \nrm{ P(X_j \gn X_i) P(X_j \gn Y_i = y') - \pi}_{TV}\\
& & \quad =\nrm{ P(X_{j-i} \gn X_1) P(X_{j-i} \gn Y_1 = y') - \pi}_{TV}\\
& & \quad \leq \sup_{x_1}\nrm{ P(X_{j-i} \gn x_1) - \pi}_{TV}
 \leq  G\theta^{i-1}.
\eeq

So calculate
\beq
&& \E[f(Y_i)f(Y_j)]= \\ & = & \int_\Y dy dy' P(Y_j = y' , Y_i = y) f(y)f(y') \\
& = & \int_\Y dy dy' P(Y_j = y' \gn Y_i = y) P(Y_i = y) f(y)f(y') \\
& \leq & \int_\Y dy dy' P(Y_i = y) \left(P(Y_j = y') + G\theta^{j-i}\right) f(y)f(y') \\
& = & \E[f(y)]^2 + G\theta^{j-i} \E[f(y)] \int_\Y f(y) dy \\
& = & \E[f(y)] \left(\E[f(y)] + G\theta^{j-i} \int_\Y f(y) dy \right).
\eeq

Thus
\beq
& & \sum_{1 \leq i < j \leq T}\E[f(Y_i)f(Y_j)]\\ 
& \leq & \sum_{1 \leq i < j \leq T} \E[f(Y)]\left(\E[f(Y)] +  G \theta^{j-i}  \right)\\
& = & {\frac{(T-1)T}{2}} \left( \E[f(Y)]\right)^2 +   G \E[f(Y)] \sum_{k=1}^T{(T-k)\theta^{k}} \\
& \leq & {\frac{(T-1)T}{2}} \left( \E[f(Y)]\right)^2 + \frac{ T G  \theta}{1-\theta}  \E[f(Y)]
\eeq
since
\beq
\sum_{k=1}^T{(T-k)\theta^{k}} = \frac{\theta}{1 - \theta}\left( T - \frac{1 - \theta^T}{1-\theta}\right) \leq  \frac{T \theta}{1 - \theta}.
\eeq
So
\beq
& &\E\left[ \left(\sum_{i=1}^T f(Y_i)\right)^2\right]\\ & = & 
 T \E[f^2(Y)] + 2 \sum_{1 \leq i < j \leq T}\E[f(Y_i)f(Y_j)]\\
 & \leq & T \E[f^2(Y)] + 2 {\frac{(T-1)T}{2}} \left( \E[f(Y)]\right)^2 + 2\frac{ T G \theta}{1-\theta}  \E[f(Y)]
\eeq
Putting it all together we get
\beq
& &\var\left[\sum_{i=1}^T f(Y_i)\right]\\ 
& = & \E\left[\left(\sum_{i=1}^T f(Y_i)\right)^2\right] - \left(\sum_{i=1}^T \E [f(Y_i)]\right)^2\\
& \leq & T \E[f^2(Y)]   - T \left( \E[f(Y)]\right)^2 + 2\frac{ T G  \theta}{1-\theta}  \E[f(Y)]\\
& = & T \left( \var[f(Y)] +   g_\theta \E[f(Y)]\right)\\
& \leq & T \left( \var[f(Y)] +  g_\theta L\right).
\eeq
Concluding
\beq
\var\left[\oo{T}\sum_{t=1}^T f(Y_t)\right] & \leq & \frac{\var[f(Y)]}{T} + \frac{ g_\theta L}{T}.
\eeq
\enpf
}

\hide{
The latter, coupled with the fact that
$\E\xi=\int_{t=0}^\infty P(\xi>t)dt$
for any non-negative random variable $\xi$, 
implies (\ref{eq:E|rhosig|}),
by taking $\xi=(\hat\rho_i-\rho_i)^2$ or $\xi=(\hat\sigma_{kk'}-\sigma_{kk'})^2$.

By stationarity, we have 
$\E\hat\bro=\bro$ and
$\E\hat\bsi=\bsi$, whereas by linearity of expectation,  
$\E\iprod{\hat\bro}{v}=\iprod{\bro}{v}$.
At this point, we invoke the following
concentration inequality for 
hidden
Markov chains
\citep{kontram06}:
if $F:(Y_0,\ldots,Y_{T-1})\mapsto \R$
is $1$-Lipschitz with respect to the 
Hamming metric on $[m]^T$, then for all
$\eps>0$,
\beqn
\label{eq:kr}
P(|F-\E F|>\eps T) \le 2 \exp(-T(1-\mix)^2\eps^2/2).
\eeqn
The observation that $\hat\rho_i$ is $(1/T)$-Lipschitz, together with the union bound,
yield the first claim.
To establish the second claim, note that 
the functional
$(Y_0,\ldots,Y_{T-1})\mapsto \iprod{\hat\bro}{v}$ is $(\sqrt2/T)$-Lipschitz.
For the third claim, observe again that the functional 
$(Y_0,\ldots,Y_{T-1})\mapsto \hat\sigma_{kk'}$ is $(T-1)\inv$-Lipschitz 
and apply the concentration inequality (\ref{eq:kr}) with the union bound.

To prove (\ref{eq:E|rho|}), recall that 
for any non-negative random variable $\xi$, $\E\xi=\int_{t=0}^\infty P(\xi>t)dt$.
Thus,
\beq
\E[ ({\hat\rho_i-\rho_i})^2 ] \!=\! \int_{0}^\infty P[ (\hat\rho_i-\rho_i)^2 > t]dt 
\le\! \int_{0}^\infty 2\exp(-\tfrac{T(1-\mix)^2 s}2)ds
= \oo{T(1-\mix)^2}.
\eeq
The proof of (\ref{eq:E|sig|}) follows from analogous arguments.
}
\end{document}